%% file: 0_main.tex
\documentclass[10pt,twocolumn,letterpaper]{article}

\usepackage[pagenumbers]{cvpr} 

\usepackage{graphicx}
\usepackage{amsmath}
\usepackage{amssymb}
\usepackage{booktabs}
\usepackage{dsfont}

\usepackage{float}
\usepackage{xcolor,colortbl}
\usepackage{subcaption}
\usepackage{optidef}
\usepackage{multirow}
\usepackage{kotex}
\usepackage{algorithmic}
\usepackage{amsfonts}
\usepackage{algorithm}

\usepackage[numbers,sort,compress]{natbib}

\usepackage[pagebackref,breaklinks,colorlinks]{hyperref}

\usepackage[capitalize]{cleveref}
\crefname{section}{Sec.}{Secs.}
\Crefname{section}{Section}{Sections}
\Crefname{table}{Table}{Tables}
\crefname{table}{Tab.}{Tabs.}

\def\cvprPaperID{11774} 
\def\confName{CVPR}
\def\confYear{2022}

\begin{document}

\title{A Highly Effective Low-Rank Compression of Deep Neural Networks \\
with Modified Beam-Search and Modified Stable Rank}

\author{Moonjung Eo\thanks{Eqaul contribution},
    Suhyun Kang\footnotemark[1],
    Wonjong Rhee
    \\
Department of Intelligence and Information, GSAI, AIIS, 
    Seoul National University\\
Seoul, 08826, South Korea\\
{\tt\small \{eod87, su\_hyun, wrhee\}@snu.ac.kr}
}
\maketitle

\begin{abstract}

Compression has emerged as one of the essential deep learning research topics, especially for the edge devices that have limited computation power and storage capacity. Among the main compression techniques, low-rank compression via matrix factorization has been known to have two problems. First, an extensive tuning is required. Second, the resulting compression performance is typically not impressive. In this work, we propose a low-rank compression method that utilizes a modified beam-search for an automatic rank selection and a modified stable rank for a compression-friendly training. The resulting $\textit{BSR}$ (Beam-search and Stable Rank) algorithm requires only a single hyperparameter to be tuned for the desired compression ratio. The performance of $\textit{BSR}$ in terms of accuracy and compression ratio trade-off curve turns out to be superior to the previously known low-rank compression methods. Furthermore, $\textit{BSR}$ can perform on par with or better than the state-of-the-art structured pruning methods. As with pruning, $\textit{BSR}$ can be easily combined with quantization for an additional compression.

\end{abstract}



\input{1_Introduction}

\input{2_related_work}

\input{3_backgrounds_and_problem_formulation}

\input{4_methodology}

\input{5_experimental_results}

\input{6_discussion}


\input{7_conclusion}

{\small
\setlength{\bibsep}{0pt}
\bibliographystyle{abbrvnat}
\bibliography{reference}
}
\clearpage
\input{8_appendix}

\end{document}

%% file: 1_Introduction.tex
\section{Introduction}
\label{sec:intro}

\begin{figure}[!t]
    \centering
    \begin{subfigure}{0.235\textwidth}
        \centering
        \includegraphics[width=\textwidth]{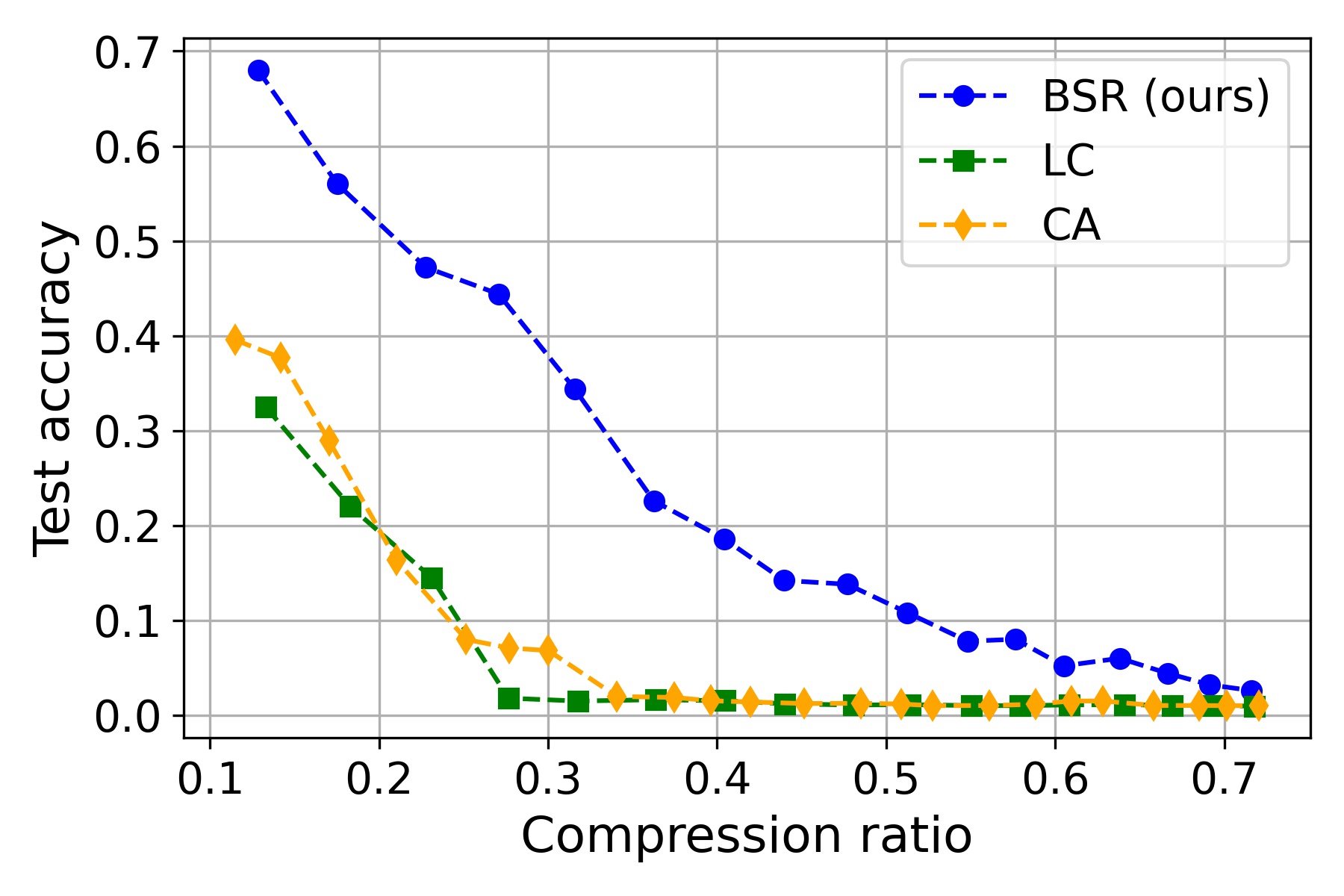}
        \caption{Rank selection}
        \label{fig:intro_a}
    \end{subfigure}
    \hfill
    \begin{subfigure}{0.235\textwidth}
        \centering
        \includegraphics[width=\textwidth]{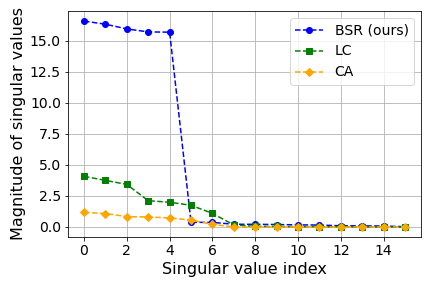}
        \caption{Singular value distribution} 
        \label{fig:intro_b}
    \end{subfigure}    
\caption{Comparison with the baseline algorithms of CA~\cite{alvarez2017compression} and LC~\cite{idelbayev2020low}. (a) Performance of rank selection methods: a base neural network (ResNet56 on CIFAR-100) was truncated by the selected ranks and no further fine-tuning was applied. Modified beam-search ($\textit{mBS}$) clearly outperforms the other two.   
(b) Effectiveness of rank regularized training: a base neural network (ResNet56 on CIFAR-100) was regularized by each compression-friendly training method. For the target rank of five, modified stable rank ($\textit{mBR}$) is the only one that clearly minimizes the singular values other than the top five.}
\label{fig:intro}
\end{figure}

As deep learning becomes widely adopted by the industry, the demand for compression techniques that are highly effective and easy-to-use is sharply increasing. The most popular compression methods include quantization~\cite{rastegari2016xnor,wu2016quantized}, pruning of redundant parameters~\cite{han2015deep,lebedev2016fast,srinivas2015data}, knowledge distillation from a large network to a small one~\cite{hinton2015distilling,kim2018paraphrasing,romero2014fitnets,zagoruyko2016paying}, and network factorization~\cite{alvarez2017compression,denton2014exploiting,masana2017domain,xue2013restructuring}. In this work, we focus on a low-rank compression that is based on a matrix factorization and low-rank approximation of weight matrices. 

A typical low-rank compression is based on a direct application of SVD (Singular Value Decomposition). For a trained neural network, the layer $l$'s weight matrix $\mathbf{W}_l$ of size $m_l \times n_l$ is decomposed and only the top $r_l$ dimensions are kept. This reduces the computation and memory requirements from $m_l n_l$ to $(m_l+n_l)r_l$ and the reduction can be large when a small $r_l$ is chosen. 

The traditional approaches can be divided into two main streams. One considers only a post-application of SVD on a fully trained neural network, and the other additionally performs a compression-friendly training before SVD.

For the first stream, the main research problem is the selection of $r_l$~\cite{jaderberg2014speeding,denton2014exploiting,tai2015convolutional, zhang2015accelerating, wen2017coordinating}. To be precise, the problem is to select $\mathbf{r}=[r_1,r_2,\cdots,r_L]^T$ for $L$ layers such that a high compression ratio can be achieved without harming the performance too much. Because the rank selection is a nonlinear optimization problem, a variety of heuristics have been studied where typically a threshold is introduced and adjusted. These works, however, failed to achieve a competitive effectiveness because a trained neural network is unlikely to maintain its performance when a small $r_l$ is selected. 
 
For the other stream, several works introduced an additional step of compression-friendly training~\cite{alvarez2017compression, idelbayev2020low, li2018constrained}.
This can be a promising strategy because it is a common sense to train a large network to secure a desirable learning dynamics~\cite{luo2017thinet, carreira2018learning}, but at the same time the network can be regularized in many different ways without harming the performance~\cite{choi2021statistical}. For rank selection, however, most of them still used heuristics and failed to achieve a competitive performance. 
Recently, \citet{idelbayev2020low}

achieved a state-of-the-art performance for low-rank compression by calculating target ranks $r$ with a variant of Eckhart-Young theorem and by performing a regularized training with respect to the weight matrices truncated according to the target rank vector $\mathbf{r}$. The existing compression-friendly works, however, require re-calculations of $\mathbf{r}$ during the training, train weight matrices that are not really low-rank, and demand for an extensive tuning especially for obtaining a compressed network of a desired compression ratio.

In this work, we enhance the existing low-rank compression methods in three different aspects. First, we adopt a modified beam-search with a performance validation for selecting the target rank vector $\mathbf{r}$. Compared to the previous works, our method allows a joint search of rank assignments over all the layers and the effect is shown in \Cref{fig:intro_a}. The previous works relied on a simple heuristic (e.g., all layers should be truncated to keep the same portion of energy \cite{alvarez2017compression}) or an implicit connection of all layers (e.g. through a common penalty parameter \cite{idelbayev2020low}). Secondly, we adopt a modified stable rank for the rank-regularized training. Because our modified stable rank does not rely on any instance of weight matrix truncation, it can be continuously enforced without any update as long as the target rank vector $\mathbf{r}$ remains constant. The previous works used a forced truncation in the middle of training~\cite{alvarez2017compression} or a norm distance regularization with the truncated weight matrices~\cite{idelbayev2020low}. In both methods, iteration between weight truncation step and training step was necessary. In our method, we calculate and set the target rank vector $\mathbf{r}$ only once. Our modified stable rank turns out to be very effective at controlling the singular values as can be seen in \Cref{fig:intro_b}. As the result of the first and the second aspects, our low-rank compression method can achieve a performance on par with or even better than the latest pruning methods. Because low-rank compression can be easily combined with quantization just like pruning, a very competitive overall compression performance can be achieved. Thirdly, our method requires only one hyperparameter to be tuned. For a desired compression ratio that is given, all that needs to be tuned is the regularization strength $\lambda$. While a low-rank compression method like LC in~\cite{idelbayev2020low} accomplished an outstanding performance, it requires an extensive tuning to identify a compressed network of a desired compression ratio. Such a difficulty on tuning can be a major drawback for the usability.  
We believe our simplification in tuning is a contribution that is as important as, or perhaps even more important than, the performance improvement.

%% file: 2_related_work.tex
\section{Related works}
\label{sec:related_work}

\subsection{DNN compression methods}

In the past decade, a tremendous progress has been made in the research field of DNN compression. Comprehensive surveys can be found in~\cite{lebedev2018speeding, cheng2017survey, deng2020model, choudhary2020comprehensive, nan2019deep}. Besides the low-rank compression discussed in Section~\ref{sec:intro}, the main algorithmic categories are quantization, pruning, and knowledge distillation. Among them, quantization and pruning are known as the most competitive compression schemes. 

Quantization reduces the number of data bits and parameter bits, and it is an actively studied area with a broad adoption in the industry~\cite{courbariaux2016binarized,gupta2015deep,han2015learning,rastegari2016xnor,wu2016quantized}. There are many flavors of quantization including binary parameterization~\cite{courbariaux2016binarized,rastegari2016xnor}, low-precision fixed-point~\cite{gupta2015deep, lin2016fixed}, and mixed-precision training~\cite{yang2021bsq, bulat2021bit}. While many of them require a dedicated hardware-level support, we limit our focus to the pure algorithmic solutions, and consider combining our method with an algorithmic quantization in Section~\ref{sec:discussion}.

Network pruning includes weight pruning~\cite{han2015deep, han2015learning} and filter pruning~\cite{chin2020towards, luo2017thinet, he2017channel, zhuang2018discrimination, he2018amc, he2018soft, liu2018rethinking, wang2020pruning}. Weight pruning selects individual weights to be pruned. Because of the unstructured selection patterns, it requires customized GPU kernels or specialized hardware~\cite{han2015deep,zhao2019efficient}. Filter pruning selects entire filters only. Because of the structured selection patterns, the resulting compression can be easily implemented with off-the-shelf CPUs/GPUs~\cite{sui2021chip}. In Section~\ref{sec:discussion}, our results are compared with the state-of-the-art filter pruning methods because both are software-only solutions that reduce the number of weight parameters.

\subsection{Rank selection}
In the past decade, a variety of rank selection methods have been studied for low-rank compression of deep neural networks. In the early studies of low-rank compression, rank selection itself was not the main focus of the research and it was performed by human through repeated experiments~\cite{jaderberg2014speeding, denton2014exploiting, tai2015convolutional}. Then, a correlation between the sum of singular values (that is called energy) and DNN performance was observed, and subsequently rank selection methods were proposed where they minimally affect the energy~\cite{zhang2015accelerating, alvarez2017compression, wen2017coordinating, li2018constrained, Yuhuirankpruning2018}. The following works formulated and solved rank selection as optimization problems where the singular values appear in the formulations~\cite{kim2019efficient, idelbayev2020low}.
In our modified beam-search method, we neither use the concept of energy nor utilize the singular values. Instead, we directly perform a search over the space of rank vectors using a validation dataset.

\subsection{Beam search}
Beam search is a technique for searching a tree, especially when the solution space is vast \cite{xu2007learning, antoniol1995language, furcy2005limited}. It is based on a heuristic of developing $K$ solutions in parallel while repeatedly inspecting adjacent nodes of the $K$ solutions in the tree. $K$ is commonly referred as beam size, and $K=1$ corresponds to the greedy search~\cite{huang2012structured} and $K=\infty$ corresponds to the breadth-first search~\cite{meister2020if}. Obviously, beam search is a compromise between the two, where $K$ is the control parameter. Beam search has been widely adopted, especially for natural language processing tasks such as speech recognition~\cite{lowerre1976speech}, neural machine translation~\cite{lowerre1976speech,boulanger2013audio}, scheduling~\cite{habenicht2002scheduling}, and
generation of a natural language adversarial attack~\cite{tengfei2021adversarial}. In our work, beam search is slightly modified to allow a search of depth-$s$ descendant nodes instead of the children nodes, and the modified beam search is employed for a joint search of the weight matrix ranks over $L$ layers, $\mathbf{r}=[r_1,r_2,\cdots,r_L]^T$.

\subsection{Stable rank and rank regularization}
Formally speaking, stable rank of a matrix is defined as the ratio between the squared Frobenius norm and the squared spectral norm~\cite{rudelson2007sampling}. Simply speaking, the definition boils down to the ratio between `sum of squared singular values' and the `squared value of the largest singular value'. Therefore, a smaller stable rank implies a relatively larger portion of energy in the largest singular value. In deep learning research, a stable rank normalization was studied for improving generalization~\cite{sanyal2019stable} and a stable rank approximation was utilized for performance tuning~\cite{choi2021statistical}. In our work, we modify the stable rank's denominator to the sum of top $r$ singular values such that we can concentrate the activation signals in the top $r$ dimensions. While our approach requires only the target rank $r$ to be specified, the previous compression-friendly training calculated the truncated matrices and directly used them for the regularization. Therefore, the previous methods required a frequent update of the truncated matrices during the training. In our $\textit{BSR}$ method, we calculate the target rank vector only once in the beginning and keep it fixed.

%% file: 3_backgrounds_and_problem_formulation.tex
\section{Low-rank compression}
\label{sec:backgrounds_and_problem_formulation}

\begin{figure*}[t]
    \centering
    \begin{subfigure}{0.65\textwidth}
        \centering
        \includegraphics[width=\textwidth]{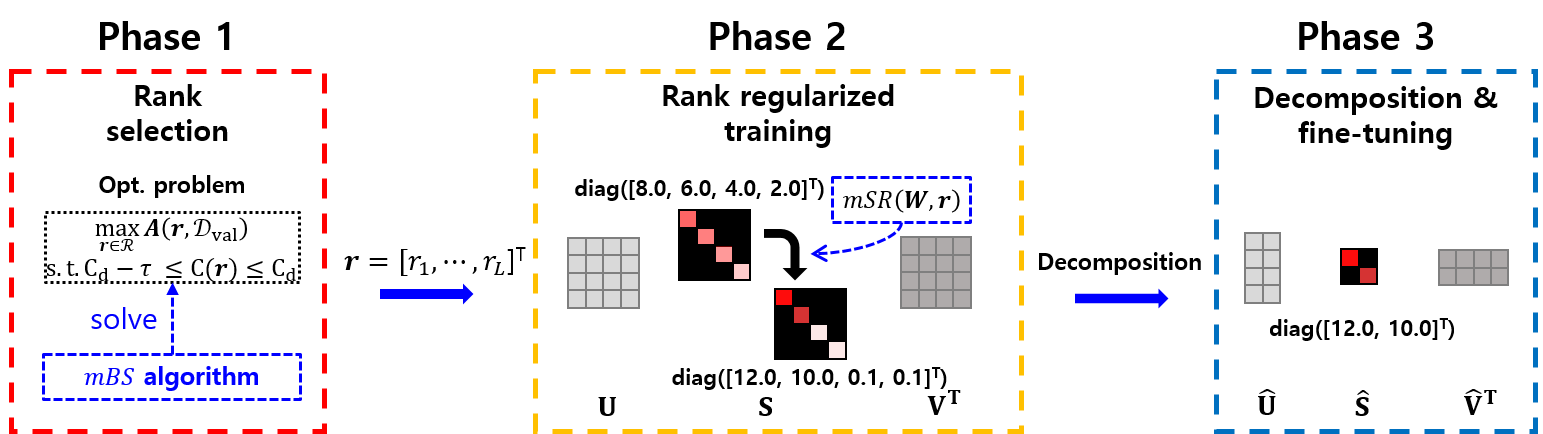}

    \end{subfigure}
    \hfill
\caption{Overall process of $\textit{BSR}$ algorithm.}
\label{fig:overall_process}
\end{figure*} 

\subsection{The basic process}
A typical process of low-rank compression consists of four steps: 1) train a deep neural network, 2) select rank assignments over $L$ layers, 3) factorize weight matrices using truncated SVD~\cite{denton2014exploiting,masana2017domain,xue2013restructuring} according to the selected ranks, and 4) fine-tune the truncated model to recover the performance as much as possible. In our work, we mainly focus on the rank selection step and an additional step of compression-friendly training. The additional step is placed between step 2 and step 3.

\subsection{Compression ratio}
Consider an $L$-layer neural network with $\mathbf{W}=(\mathbf{W}_1,\mathbf{W}_2,\cdots,\mathbf{W}_L)$, where $\mathbf{W}_l \in \mathbb{R}^{m_l\times n_l}$, as its weight matrices. Without a low-rank compression, the rank vector $\mathbf{r}$ corresponds to the full rank vector of $\mathbf{r}_{full}=[R_1,\cdots,R_L]^T$ where $R_l=\min(m_l,n_l)$. The rank selection is performed over the set of $\mathcal{R}=\{\mathbf{r}\in \mathbb{N}^L\; | \; \mathbf{r}=[r_1,r_2,\cdots,r_L]^T, 0 < r_l \leq \min(m_l,n_l) \}$, and the selected rank vector is denoted as $\mathbf{r}_{select}$. With $\mathbf{r}_{select}$, we can perform a truncated SVD. For $l$th layer's weight matrix $\mathbf{W}_l = \mathbf{U}_l\mathbf{S}_l\mathbf{V}_l^T$, we keep only the largest $r_{l}$ singular values to obtain $\mathbf{\hat{W}}_l = \mathbf{\hat{U}}_l\mathbf{\hat{S}}_l\mathbf{\hat{V}}_l^T$ where $\mathbf{\hat{U}}_l\in \mathbb{R}^{m \times r_{l}}$, $\mathbf{\hat{S}}_{l}\in\mathbb{R}^{r_{l} \times r_{l}}$, and $\mathbf{\hat{V}}_l\in \mathbb{R}^{n \times r_{l}}$. Then, the compression is achieved by replacing $\mathbf{W}_l$ with a cascade of two matrices: $\mathbf{\hat{S}}_l\mathbf{\hat{V}}_{l}^{T}$ and $\mathbf{\hat{U}}_l$. Obviously, the computational benefit stems from the reduction in the matrix multiplication loads: $mn$ for the original $\mathbf{W}_l$ and $(m+n)r_{l}$ for $\mathbf{\hat{S}}_{l}\mathbf{\hat{V}}_{l}^{T}$ and $\mathbf{\hat{U}}_l$. Similar results hold for convolutional layers. Finally, the compression ratio $\textit{C}(\mathbf{r})$ for the selected rank vector \textbf{$\mathbf{r}$} can be calculated as
\begin{equation}
    \label{eqn:compression_rate}
    \textit{C}(\mathbf{r})= 1 - \frac{\sum_{l=1}^{L} \{ m_ln_l\mathds{1}_l + r_l(m_l+n_l)(1-\mathds{1}_l) \} }{\sum_{l=1}^{L}m_ln_l},
\end{equation}
where $\mathds{1}_l$ is a simplified notation of $\mathds{1}(m_l,n_l,r_l)$ that is defined as $1$ if 
$m_ln_l \leq r_l(m_l+n_l)$ and $0$ otherwise.

%% file: 4_methodology.tex
\section{Methodology}
\label{sec:methodology}
\begin{figure}[!t]
    \centering
    \begin{subfigure}{0.42\textwidth}
        \centering
        \includegraphics[width=\textwidth]{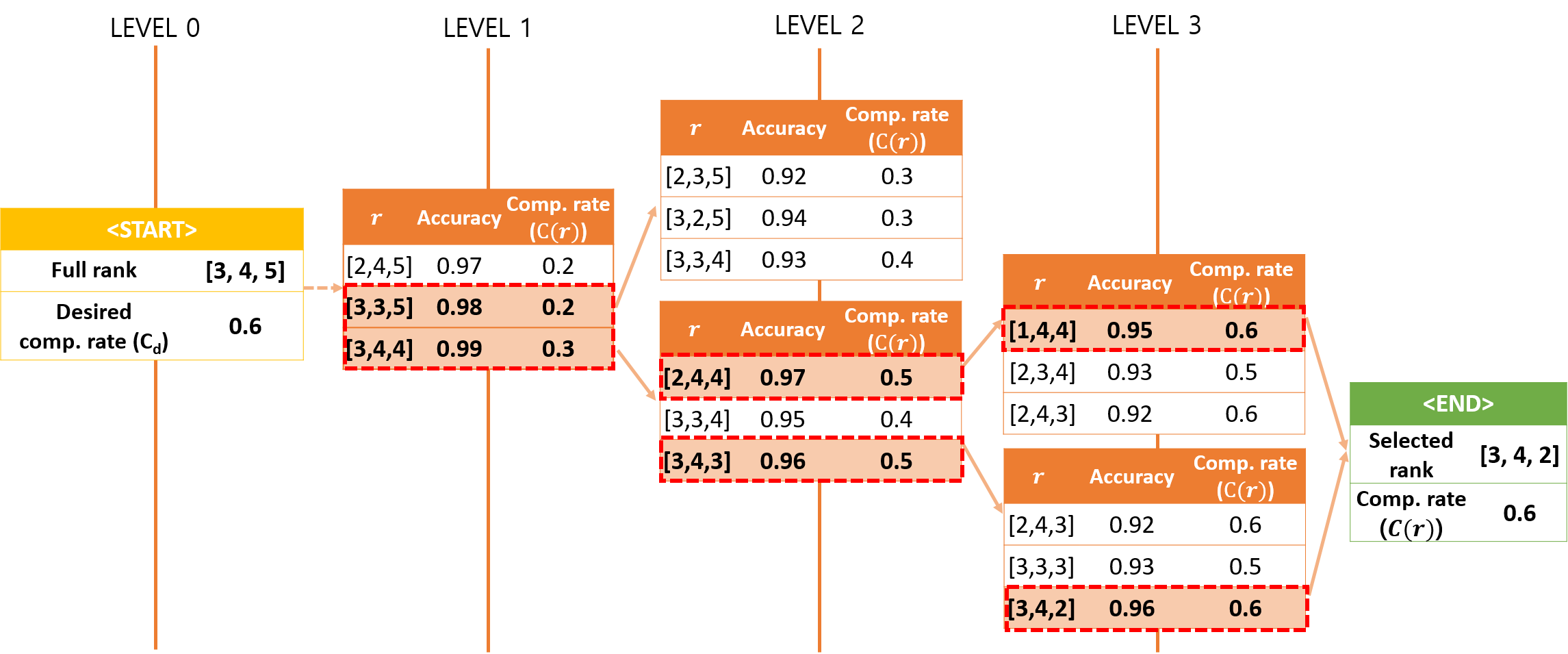}
    \end{subfigure}
    \hfill
\caption{Illustration of $\textit{mBS}$ search process for $L=3$, $K=2$, $s=1$, and $C_d=0.6$.}
\label{fig:selection_algorithm}
\end{figure}

\subsection{Overall process}
The overall process of our $\textit{BSR}$ low-rank compression is shown in \Cref{fig:overall_process}. Starting from a fully trained network, the phase one of $\textit{BSR}$ performs rank selection using $\textit{mBS}$ algorithm where it requires only a desired compression ratio as the input. Once the phase one is completed,  $\mathbf{r}_{select}$ is fixed and rank regularized training is performed in phase two. The strength of $\textit{mSR}$ is controlled by $\lambda$ where its strength is gradually increased in a scheduled manner. Upon the completion of the phase two, the trained network is truncated using singular value decomposition according to $\mathbf{r}_{select}$. Then, a final fine-tuning is performed to complete the compression.

\subsection{Modified beam-search ($\textit{mBS}$) for rank selection}
When a neural network with $\mathbf{W}$ is truncated according to the rank vector $\mathbf{r}$, the accuracy can be evaluated with a validation dataset $\mathcal{D}_{val}$ and the accuracy is denoted as $\textit{A}(\mathbf{r}, \mathcal{D}_{val})$. The corresponding compression ratio can be calculated as $\textit{C}(\mathbf{r})$.
Our goal of rank selection is to find the $\mathbf{r}$ with the highest accuracy while the compression ratio is sufficiently close to the desired compression ratio $C_d$. This problem can be formulated as below.
\begin{equation}
    \label{eqn:combinatorial}
    \begin{aligned}
       & \max\limits_{\mathbf{r}\in\mathcal{R}}\textit{A}(\mathbf{r}, \mathcal{D}_{val})\\
       & \textrm{s.t.}\quad C_d-\tau\leq\textit{C}(\mathbf{r}) \leq C_d
    \end{aligned}
\end{equation}
Note that we have introduced a small constant $\tau$ for relaxing the desired compression ratio. This relaxation forces the returned solution to have a compression ratio close to $C_d$, and it is also utilized as a part of the exit criteria.  

The problem in \Cref{eqn:combinatorial} is a combinatorial optimization problem~\cite{reeves@1993comb} and a simple greedy algorithm can be applied as in~\cite{zhang2015accelerating}. 
Because the cardinality of the search space $\mathcal{R}$ is  extremely large, however, greedy algorithms hardly produce good results in a reasonable computation time. On the other hand, a full search is also unacceptable because of its long search time. As a compromise, we adopt a beam-search framework and make adequate adjustments. Before presenting the details of $\textit{mBS}$, a simple illustration of how \Cref{eqn:combinatorial} can be solved with our modified beam search is presented in \Cref{fig:selection_algorithm}.

The details of $\textit{mBS}$ can be explained as the following.
\begin{itemize}
    \item \textbf{Stage 1:} Initialize level, $lv_{[1]}=1$. Initialize top-K set, $\mathcal{B}_{[1]}=\{\mathbf{r}^1_{[1]}\}$, where $\mathbf{r}^1_{[1]}=\mathbf{r}_{full}$ (full rank assignments). 
    \item \textbf{Stage 2:} Move to the next level by adding the pre-chosen step size $s$, $lv_{[t]} = lv_{[t-1]}+s$. 
    For each element in $\mathcal{B}_{[t-1]}$, find all of its descendants in $lv_{[t]}$ and add them to candidate set $\mathcal{T}_{[t]}$. Exclude the descendants that performs too much compression by checking the condition  $\textit{C}(\mathbf{r}_{[t]})\leq C_d$.

    \item \textbf{Stage 3:} Calculate the new top-K set at level $lv_{[t]}$ by finding the ordered top $K$ elements: 
    \begin{equation*}
        \mathcal{B}_{[t]}=\underset{\mathbf{r}^1_{[t]},\cdots\mathbf{r}^K_{[t]}\in\mathcal{T}_{[t]}}{\arg\max}\textit{A}(\mathbf{r}_{[t]}, \mathcal{D}_{val})
    \end{equation*}
    \item \textbf{Stage 4:} Repeat \textbf{Stage 2} and \textbf{Stage 3} until $C_d-\tau \leq \textit{C}(\mathbf{r}^{1}_{[t]}) \leq C_d$ is satisfied for the best element $\mathbf{r}_{[t]}^{1}$. If the condition is satisfied, return $\mathbf{r}^{1}_{[t]}$ as $\mathbf{r}_{select}$.
\end{itemize}
An implementation of this process is provided in \Cref{alg:rank_selection_alg}. The main modification we make is the introduction of the \textit{level step size} $s$. For rank selection of low-rank compression, a weight matrix's rank needs to be reduced by a sufficient amount and meets the minimum condition of $r_l(m_l+n_l) \leq m_ln_l$ to achieve a positive compression effect. Furthermore, most of the weight matrices can allow a significant amount of rank reduction without harming the performance in the beginning of the search. Therefore, we typically set $s$ between three and ten to make the search faster. As the search progresses, we reduce $s$ whenever no candidate can be found at the next level and improve the search resolution. Besides $s$, the beam size $K$ is another important parameter that determines the trade-off between speed and search resolution. Instead of performing a fine and slow search with a small $s$ and a large $K$, we perform a fast search a few times with different configurations and choose the best $\mathbf{r}_{select}$. The configuration details can be found in \Cref{sec:Experiments} and ablation studies for $K$ and $s$ can be found in \Cref{sec:discussion}.

\begin{algorithm}[!t]
    \footnotesize
    \caption{modified Beam Search ($\textit{mBS}$) for rank selection}\label{alg:rank_selection_alg}
    \textbf{Input}: desired compression ratio $\textit{C}_d$; validation data $\mathcal{D}_{val}$; beam size $K$; level step size $s$\\ 
    \textbf{Output}: selected rank $\mathbf{r}_{select}$\\
    \textbf{Required}: ratio function $\textit{C}(\mathbf{r})$; base network $\mathbf{M}(\mathbf{r})$ with rank $\mathbf{r}$; evaluation function $\textit{A}(\mathbf{r}, \mathcal{D})$\\
    \textbf{Initialize}: $\mathbf{r}_{select}\leftarrow\mathbf{r}_{full}$;\\
    top-$K$ rank set $\mathcal{B} \leftarrow \{(\mathbf{r}_{full},\;\textit{C}(\mathbf{r}_{full}),\;\textit{A}(\mathbf{r}_{full},\mathcal{D}_{val})\}$
    \begin{algorithmic}[1]
        \WHILE{($\mathcal{B}$ is changed) $\vee$ ($C_d-\tau\leq\mathbf{r}_{select}\leq C_d$)}
            \STATE $\mathcal{T}=\{\phi\}$
            \FOR{$(\mathbf{r}_p,\;C_p,\;A_p)$ in $\mathcal{B}$}
                \FOR{$i=1$ to $L$}
                        \STATE $\mathbf{r}_c \leftarrow \mathbf{r}_p$
                        \STATE $\mathbf{r}_c[i] \leftarrow \mathbf{r}_c[i] - s$
                        \STATE $C_c \leftarrow \textit{C}(\mathbf{r}_c)$
                    \IF{$C_c \leq C_d$}
                        \STATE $A_c \leftarrow \textit{A}(\mathbf{r}_c, \mathcal{D}_{val})$ 
                        \STATE $\mathcal{T} \cup (\mathbf{r}_{c},C_c,A_c)$
                    \ENDIF
                \ENDFOR    
            \ENDFOR
            \STATE $\mathcal{B} \leftarrow \mathbf{TopK}(\mathcal{T};keys=[\mathbf{A},\mathbf{r}, \mathbf{rand}])$
            \STATE $\mathbf{r}_{select}\leftarrow \mathcal{B}[0]$
         \ENDWHILE\\
      \STATE \textbf{return} $\mathbf{r}_{select}$ 
    \end{algorithmic}
\end{algorithm}

\subsection{Modified stable rank ($\textit{mSR}$) for regularized training} 
\label{sec:msr}
For a weight matrix $ \mathbf{W}_l \in \mathbb{R}^{m_l\times n_l}$, stable rank is defined as 
\begin{equation}
    \begin{aligned}
        \label{eqn:stable_rank}
            && \textit{SR}(\mathbf{W}_l) = \frac{\left\| \mathbf{W}_{l} \right\|^{2}_{F}}{\left\| \mathbf{W}_{l} \right\|^{2}_{2}} = \frac{\sum_{i=1}^{R_l}(\sigma^{l}_{i})^2}{(\sigma^{l}_{i})^2},
    \end{aligned}
\end{equation}
where $\sigma^{l}_{i}$ is the $i$th singular value of $\mathbf{W}_l$. Because our goal of compression-friendly training is to have almost no energy in the dimensions other than the top $r_l$ dimensions, we modify the stable rank as below. 

\begin{equation}
\begin{aligned}
    \label{eqn:rank_regularizer}
    && \textit{mSR}(\mathbf{W}_l, r_l) = \frac{tr(\mathbf{\Sigma}_l^{r_l:R_l})}{tr(\mathbf{\Sigma}_l^{1:r_l})} = \frac{\sum_{i=r_l+1}^{R_l}\sigma^l_{i}}{\sum_{i=1}^{r_l}\sigma^l_{i}} \\
\end{aligned}
\end{equation}

The modified stable rank $\textit{mSR}$ is different from the stable rank in four ways. First, it is dependent to the input parameter $r_l$. Second, the summation in the denominator is performed over the largest $r_l$ singular values. Third, the largest $r_l$ singular values are excluded in the numerator's summation. Fourth, the singular values are not squared. The third and fourth differences make $\textit{mSR}$ regularization result less energy in the undesired dimensions as shown in \Cref{fig:intro_a}. The compression-friendly training is performed by minimizing the loss of  $\mathcal{L}(\mathbf{W}) + \lambda \sum_{l=1}^L\textit{mSR}(\mathbf{W}_l,r_l)$, where the first term is the original loss of the learning task and the second term is the $\textit{mSR}$ as a penalty loss. 

Through our empirical evaluations, we have confirmed that $\textit{mSR}$ can stably affect the weight matrices. In fact, the gradient of $\textit{mSR}$ can be easily derived. To do so, we decompose $\mathbf{W}_l = \mathbf{U}_l\mathbf{\Sigma }_l\mathbf{V}_l^{T}$ into two parts by allocating the first $r_l$ dimensions into $\mathbf{W}_l^{1:r_l}$ and the remaining dimensions into $\mathbf{W}_l^{r_l:R_l}$ as below.  
\begin{align*}
\small
\begin{split}
    \mathbf{W}_l &= \mathbf{W}_l^{1:r_l} \mathbf{W}_l^{r_l:R_l} \\
            &  = \mathbf{U}_l^{1:r_l} \mathbf{\Sigma}_{l}^{1:r_l} (\mathbf{V}_l^{1:r_l})^{T} + \mathbf{U}_l^{r_l:R_l} \mathbf{\Sigma}_{l}^{r_l:R_l} (\mathbf{V}_l^{r_l:R_l})^{T}
\end{split}
\end{align*}

Then, the derivative can be derived as the following. The details of the derivation are deferred to Appendix A.
\begin{align*}
\footnotesize
    \begin{split}
        & \qquad \frac{\partial \textit{mSR}(\mathbf{W}_l,r_l)}{\partial \mathbf{W}_l} = \frac{\partial \left(\frac{tr(\mathbf{\Sigma}_l^{r_l:R_l})}{tr(\mathbf{\Sigma}_l^{1:r_l})}\right)}{\partial \mathbf{W}_l} \\
        & = \frac{1}{(tr(\mathbf{\Sigma}_l^{1:r_l}))^{2}}\left(tr(\mathbf{\Sigma}_l^{1:r_l})\frac{\partial tr(\mathbf{\Sigma}_l^{r_l:R_l})}{\partial \mathbf{W}_l}-tr(\mathbf{\Sigma}_l^{r_l:R_l})\frac{\partial tr(\mathbf{\Sigma}_l^{1:r_l})}{\partial \mathbf{W}_l}\right) \\
        & = \frac{tr(\mathbf{\Sigma}_{l}^{r_l:R_l})}{tr(\mathbf{\Sigma}_{l}^{1:r_l})}\left(\frac{\mathbf{U}_{l}^{r_l:R_l}(\mathbf{V}_{l}^{r_l:R_l})^{T}}{tr(\mathbf{\Sigma}_{l}^{r_l:R_l})} - \frac{\mathbf{U}_{l}^{1:r_l}(\mathbf{V}_{l}^{1:r_l})^{T}}{tr(\mathbf{\Sigma}_{l}^{1:r_l})}\right). 
    \end{split}
\label{eqn:derivative}
\end{align*}

A crucial issue with the above $\textit{mSR}$ regularization is its effect on the computational overhead. Use of $\textit{mSR}$ requires a repeated calculation of singular value decomposition (SVD) that is computationally intensive. To deal with this issue, we adopted the randomized matrix decomposition method in \cite{erichson2016randomized}. Because the target rank $r_l$ is typically chosen to be small, the practical choice of implementation has almost no effect on the regularization while significantly reducing the computational burden. Furthermore, we obtain an additional reduction by calculating SVD only once every 64 iterations. Consequently, the training time of $\textit{mSR}$ regularization remains almost the same as the un-regularized training.

%% file: 5_experimental_results.tex
\section{Experiments}
\label{sec:Experiments}

\subsection{Experimental setting}

\subsubsection{Baseline models and datasets} 
To investigate the effectiveness and generalizability of $\textit{BSR}$, we evaluate its compression performance for a variety of models and datasets. We mainly followed the experimental settings of LC~\cite{idelbayev2020low}: LeNet5 on MNIST, ResNet32, and ResNet56 on CIFAR-10, ResNet56 on CIFAR-100, and AlexNet on large-scale ImageNet~(ILSVRC 2012). 

\subsubsection{Rank selection configuration} 
BSR performs rank selection only once. Therefore, it is important to find a rank vector that can result in a high-performance after compression-friendly training. To improve the search speed and to make the search algorithm \textit{mBS} robust, we use three settings of $(s, K)$ as $\{(3,5), (5,5), (10,5)\}$. Search for $\mathbf{r}_{select}$ is performed three times with the three settings, and the best performing one is selected as the final solution. Compression-friendly training is performed only after the final selection. Because we choose $s$ that is larger than one, \textit{mBS} might fail to find a solution. When this happens, the level step size $s$ is multiplied by $\gamma=0.5$ and the search is continued from the last $K$ candidates.

\subsubsection{Rank regularized training configuration}
\label{sec:5.1.3}
We have considered $\{0.1,\;0.01\}$ for the initial learning rate $\eta_0$ and a cosine annealing method was used as the learning rate scheduler. We used Nesterov’s accelerated gradient method with momentum 0.9 on mini-batches of size 128. 
A regularization strength scheduling was introduced for a stable regularized training, where $\lambda$ was gradually increased. To be specific, we used a regularization strength scheduling of $\lambda_{j}$ = $\lambda_{0}$·b$_{j}$ with $\lambda_{0}$ as either $2·10^{-2}$ or $5·10^{-2}$ and b$_{j}\in \{1.2,\;1.5\}$ for every 15 epochs. 

\subsection{Experimental results}
We have evaluated the performance in terms of compression ratio vs. test accuracy, and the results are provided here. We have additionally evaluated FLOPs vs. test accuracy, and the results can be found in the appendix.

\subsubsection{MNIST}

\begin{figure}[!t]
    \centering
    \includegraphics[width=0.3\textwidth]{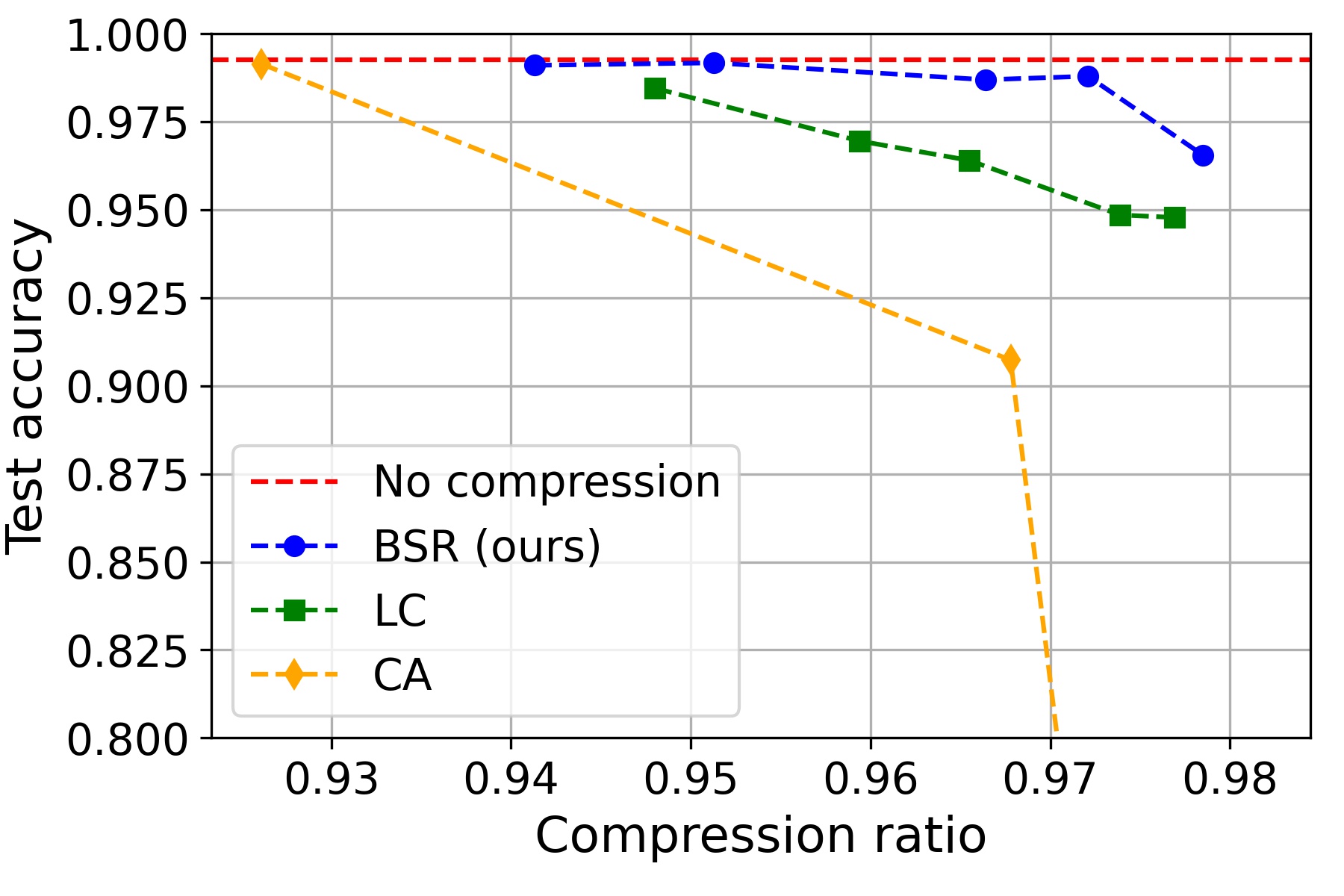}
    \caption{Comparison of $\textit{BSR}$ with CA and LC for LeNet5 on MNIST.}
    \label{fig:LeNet-5 on MNIST}
\end{figure}

The results for LeNet5 on MNIST are shown in \Cref{fig:LeNet-5 on MNIST}. 

Both of $\textit{BSR}$ and LC clearly outperform CA. Between BSR and LC, BSR outperforms LC for the entire evaluation range. It can be seen that BSR's test accuracy is hardly reduced until the compression ratio becomes 0.97. 

\subsubsection{CIFAR-10 and CIFAR-100}
 
\begin{figure*}
     \centering
     \begin{subfigure}[b]{0.3\textwidth}
         \centering
         \includegraphics[width=\textwidth]{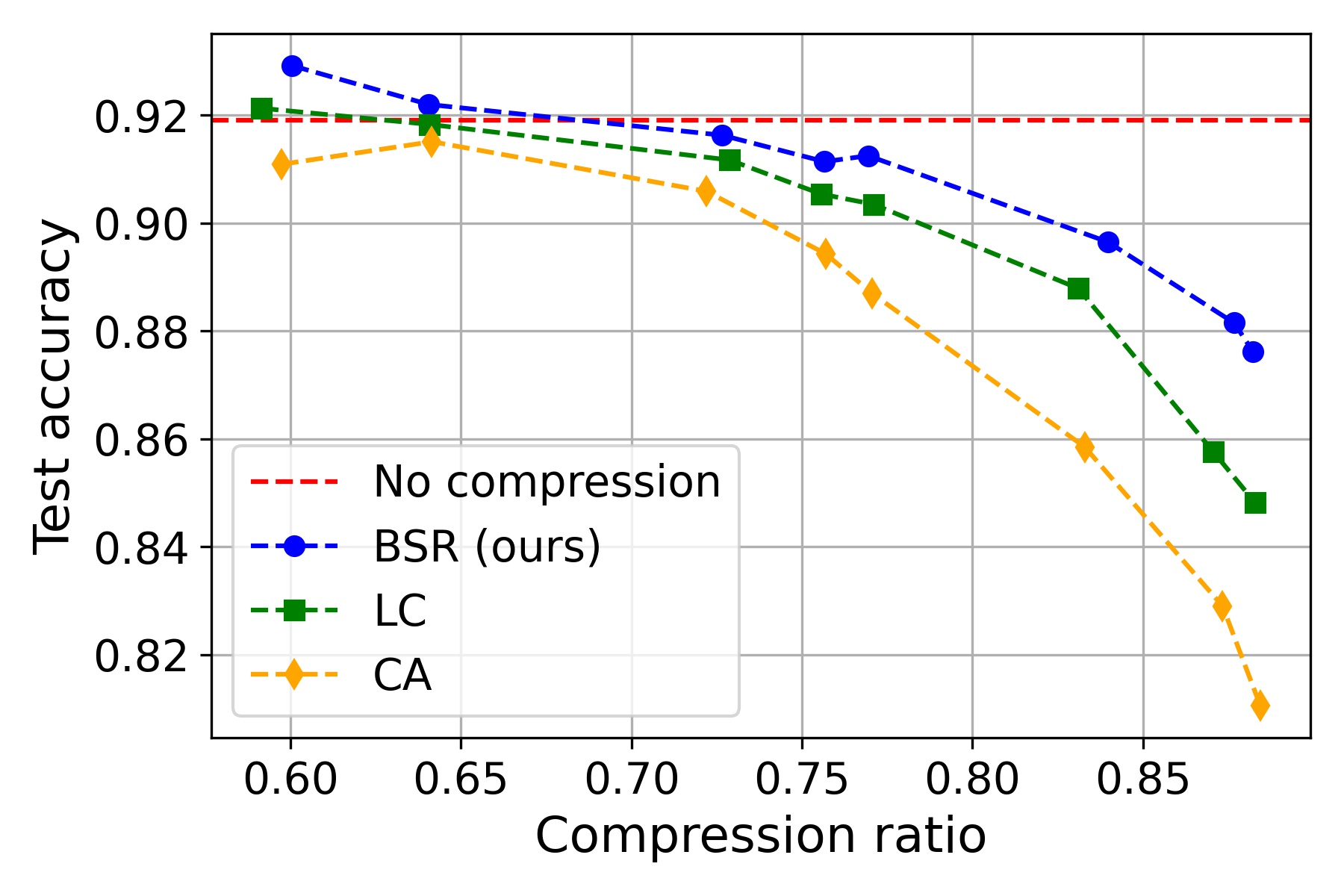}
         \caption{ResNet32 on CIFAR-10}
         \label{fig:ResNet32 on CIFAR10}
     \end{subfigure}
     \hfill
     \begin{subfigure}[b]{0.3\textwidth}
         \centering
         \includegraphics[width=\textwidth]{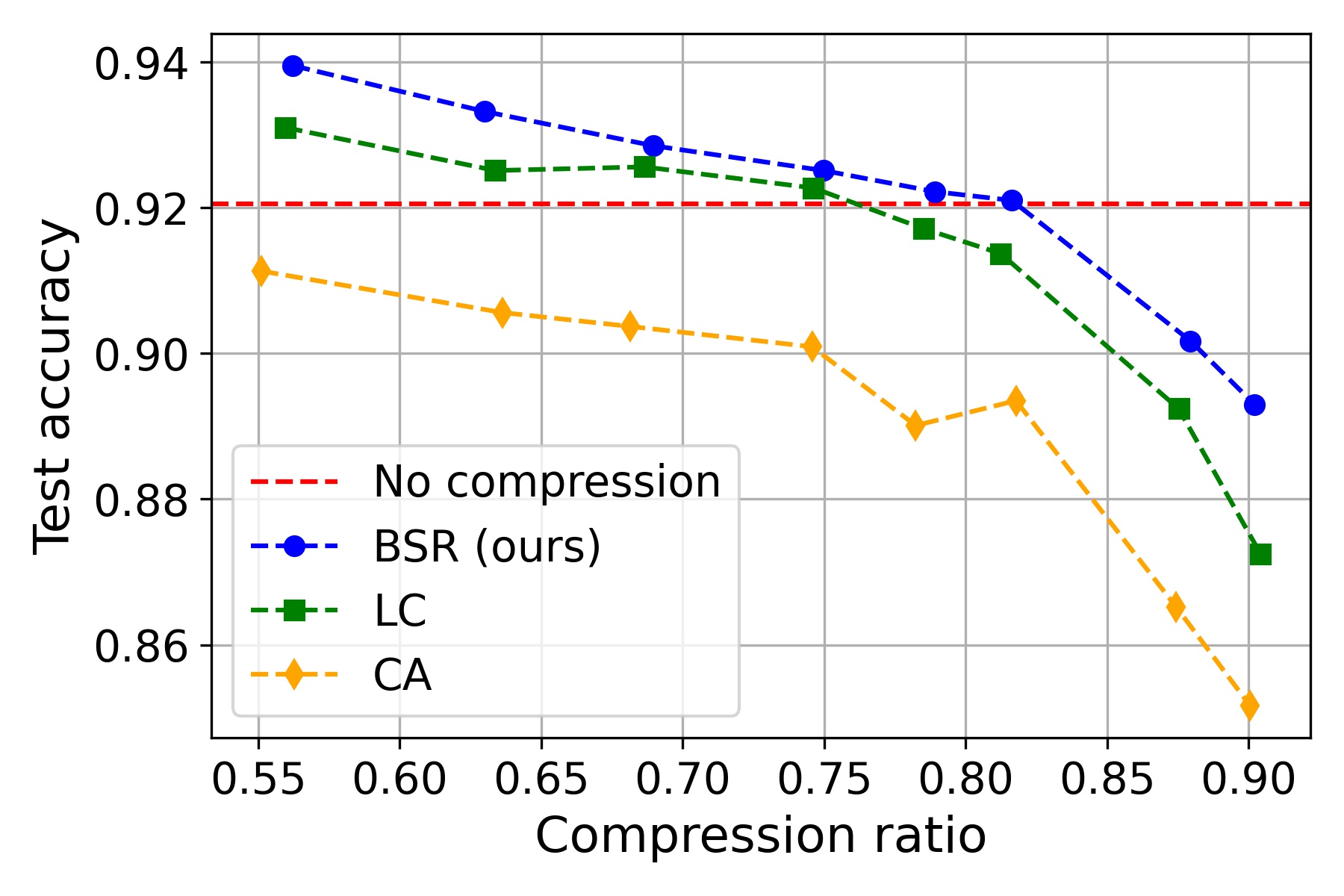}
         \caption{ResNet56 on CIFAR-10}
         \label{fig:ResNet56 on CIFAR10}
     \end{subfigure}
      \hfill
     \begin{subfigure}[b]{0.3\textwidth}
         \centering
         \includegraphics[width=\textwidth]{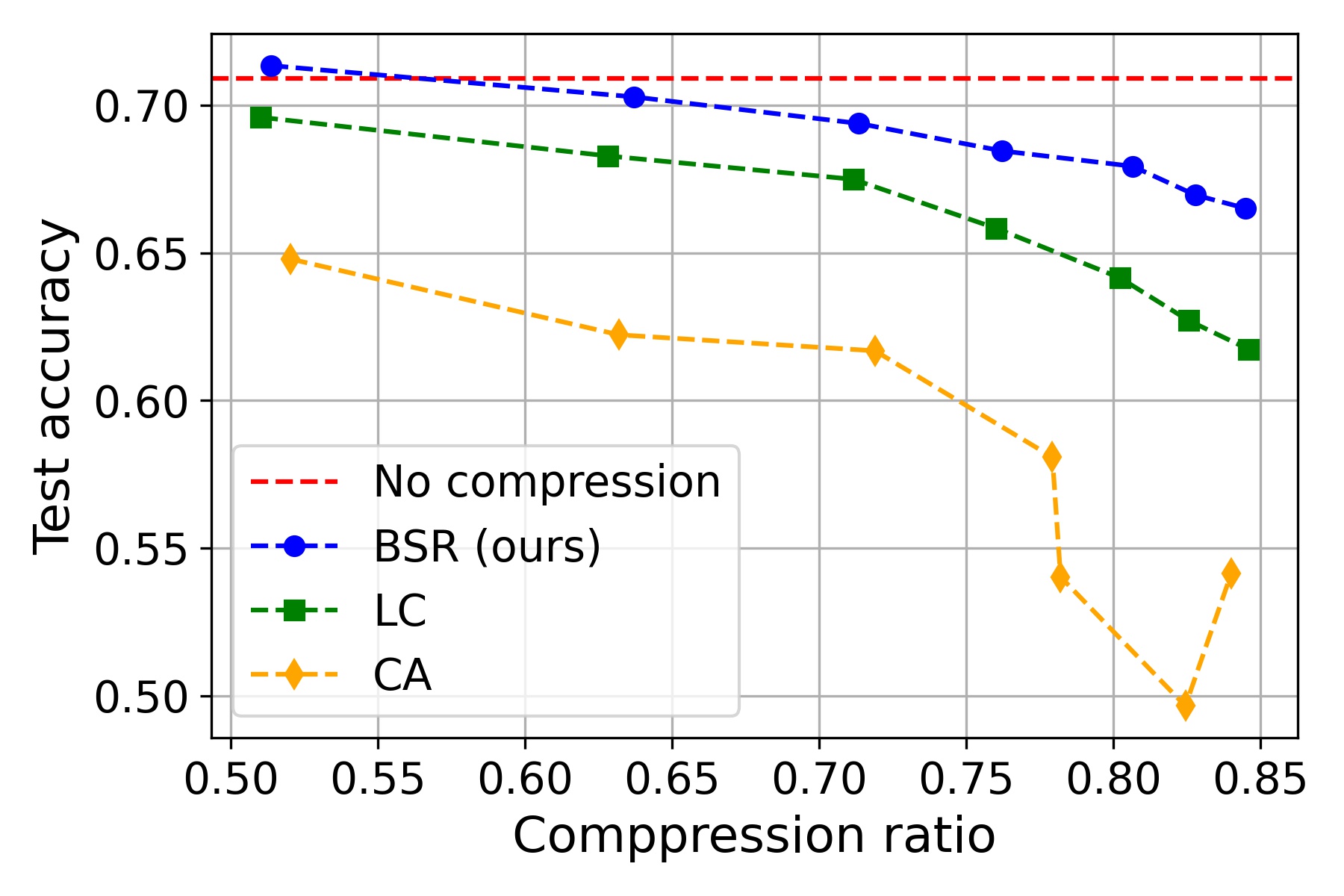}
         \caption{ResNet56 on CIFAR-100}
         \label{fig:ResNet56 on CIFAR100}
     \end{subfigure}

\caption{Comparison of $\textit{BSR}$ with CA and LC for (a) ResNet32 on CIFAR-10, (b) ResNet56 on CIFAR-10, and (c) ResNet56 on CIFAR-100.}
\label{fig:ResNet on CIFAR}
\end{figure*}

The results for ResNet32 and ResNet56 on CIFAR-10 and CIFAR-100 are shown in \Cref{fig:ResNet on CIFAR}. For ResNet, \textit{BSR} performs a low-rank compression over the convolutional layers and achieves a conspicuous improvement in compression performance. It can be also noted that $\textit{BSR}$ can even improve the no-compression accuracy when compression ratio is not too large. For instance, test accuracy in  \Cref{fig:ResNet56 on CIFAR10} is improved from 0.92 to 0.94 by \textit{BSR} when the compression ratio is around 0.55. LC is also able to improve the accuracy, but the gain is smaller than \textit{BSR}. This implies that the regularized training methods can have a positive influence on learning dynamics. As in MNIST, the performance gap between $\textit{BSR}$ and the other methods becomes larger as the compression ratio increases. 

For CA and LC, compression ratio cannot be fully controlled and the final compression ratio is dependent on the hyperparameter setting. We have tried many different hyperparameter settings to generate the CA and LC results. For BSR, the compression ratio can be fully controlled because it is an input parameter. Therefore, we have first generated LC results and have chosen the compression ratios of BSR evaluations to be the same as what LC ended up with.

\subsubsection{ImageNet}

\begin{figure}[!t]
    \centering
    \includegraphics[width=0.3\textwidth]{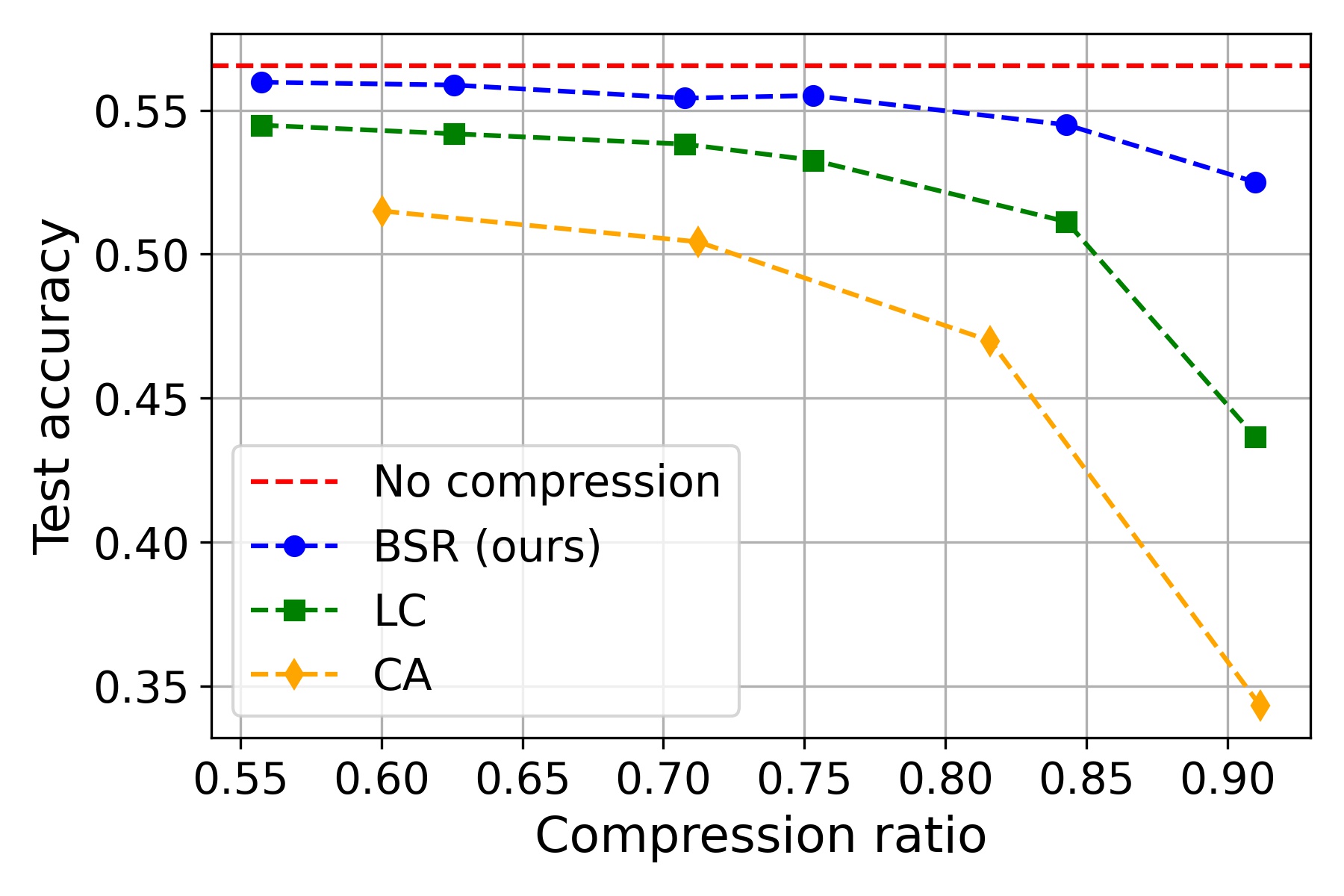}
    \caption{Comparison of $\textit{BSR}$ with CA and LC for AlexNet on ImageNet (Top-1 performance).}
    \label{fig:Alexnet on ImageNet}
\end{figure}

The results for AlexNet on ImageNet are shown in \Cref{fig:Alexnet on ImageNet}. 
We used the pre-trained Pytorch AlexNet network as the base network. The network achieves accuracy of 56.55\% for top-1 and 79.19\% for top-5. ImageNet is a much more realistic dataset than MNIST or CIFAR. The performance curves show similar patterns as in \Cref{fig:LeNet-5 on MNIST} and \Cref{fig:ResNet on CIFAR}, confirming that \textit{BSR} works well for realistic datasets. 
         
%% file: 6_discussion.tex
\section{Discussion}
\label{sec:discussion}
\subsection{Ablation test}

\begin{figure}[!ht]
    \centering
    \begin{subfigure}{0.235\textwidth}
        \centering
        \includegraphics[width=\textwidth]{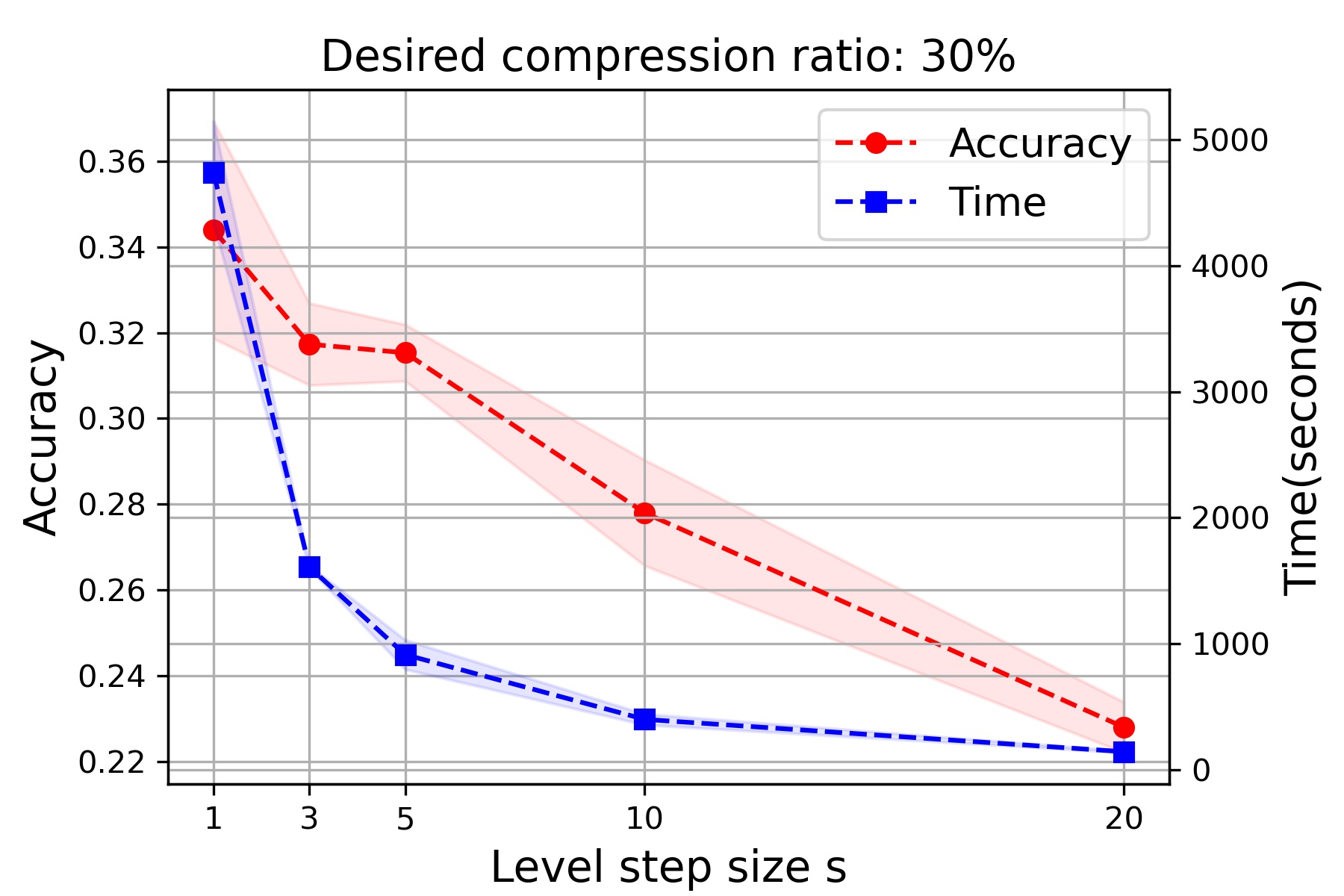}
        \caption{Effect of level step size $s$} 
        \label{subfig:k=1}
    \end{subfigure}
    \hfill
    \begin{subfigure}{0.235\textwidth}
        \centering
        \includegraphics[width=\textwidth]{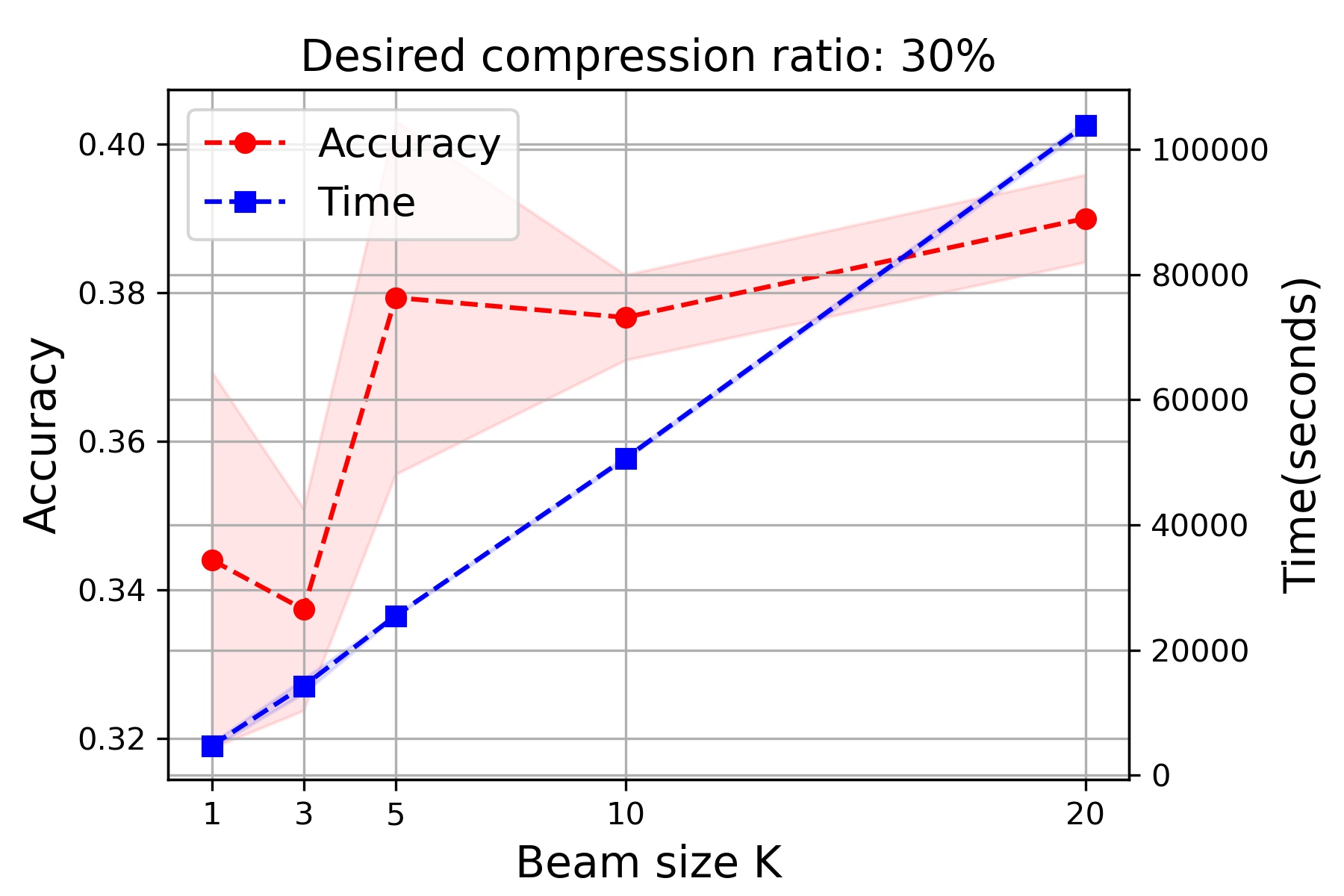}
        \caption{Effect of beam size $K$}
        \label{subfig:init=1}
    \end{subfigure}
\caption{The effect of $K$ and $s$ parameters on $\textit{mBS}$'s performance: a base neural network (ResNet56 on CIFAR-100) was truncated by the selected ranks and no further fine-tuning was applied for this analysis. Performance and search speed change (a) as $s$ increases when $K$ is fixed to 1 (b) as $K$ increases when $s$ is fixed to 1.}
\label{fig:m, initial step}
\end{figure}

\paragraph{The effect of level step size $s$:}
We explore the effect of $s$ on the performance of $\textit{mBS}$ by evaluating the selected rank's quality as a function of $s$ for $C_d=0.3$, $K=1$, and $\gamma=0.5$. The results are shown in \Cref{subfig:k=1}, and we can observe that a smaller $s$ provides a better accuracy performance. On the contrary, a smaller $s$ makes the search time exponentially larger, especially for a very small $s$. Because of the trade-off, we use $s$ between three and ten with an adaptive reduction of $s$ when no solution is found. 

\paragraph{The effect of beam size $K$:}
We explore the effect of $K$ on the performance of $\textit{mBS}$ by evaluating the selected rank's quality as a function of $K$ for $C_d=0.3$, $s=1$, and $\gamma=0.5$.
The results are shown in \Cref{subfig:init=1}, and we can observe that a larger $K$ provides a better accuracy performance in general. The accuracy curve, however, exhibits a large variance and the average performance is even deteriorated when $K$ is increased from one to three. This can be attributed to the nature of the rank selection problem. Because it is a non-convex problem, it is difficult to say what to expect. Compared to the accuracy curve, the search time curve shows a monotonic behavior where search time increases as $K$ is increased. Based on the results in \Cref{subfig:init=1}, we have chosen $K$ to be five.  

\begin{figure}[!t]
\begin{center}
   
    \begin{subfigure}{0.235\textwidth}
       
        \includegraphics[width=\textwidth]{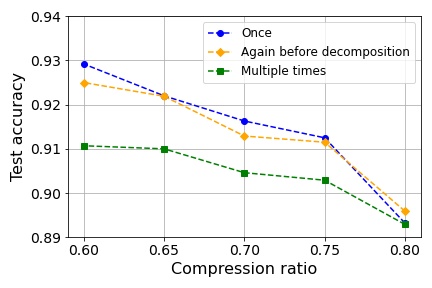}
        \caption{Rank selection updates} 
        \label{subfig:rs}
    \end{subfigure}
    \hfill
    \begin{subfigure}{0.235\textwidth}
       
        \includegraphics[width=\textwidth]{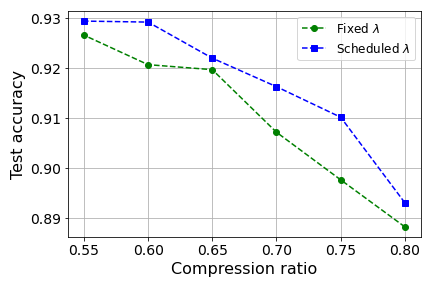}
        \caption{Scheduled increase of $\lambda$}
        \label{subfig:lambda}
    \end{subfigure}
\caption{Performance of \textit{BSR} for ResNet32 on CIFAR-10: (a) For rank selection, the best performance is achieved when the rank vector is set once and not updated. For multi-time, we have updated the target rank vector at every 30 epochs. (b) For the scheduling of the strength of $\lambda$, the performance is improved with a scheduled increase in strength.} 
\label{fig:ablation}
\end{center}
\end{figure}

\paragraph{Update of $\mathbf{r}_{select}$ during training:} In the previous works of CA and LC, the rank vector $\mathbf{r}$ is a moving target in the sense that CA performs truncated SVD multiple times during the compression-friendly training and LC updates the target weight matrices multiple times during the compression-friendly training. To investigate if \textit{BSR} can benefit by updating $\mathbf{r}_{select}$, we have compared three different scenarios - $\mathbf{r}_{select}$ is calculated only once in the beginning (``once"), $\mathbf{r}_{select}$ is additionally updated once just before the decomposition and final fine-tuning (i.e., just before the phase three in \Cref{fig:overall_process}; ``again before decomposition"), and $\mathbf{r}_{select}$ is updated at every 30 epochs (``multiple times"). The results are shown in \Cref{subfig:rs}. Interestingly, \textit{BSA} performs best when $\mathbf{r}_{select}$ is determined only once in the beginning and never changed until the completion of the compression. This is closely related to the characteristics of \textit{mSR}. As already mentioned, regularization through the modified stable rank does not rely on any particular instance of weight matrices. In fact, it only needs to know the target rank vector to be effective. Therefore, there is no need for any update during the training, and \textit{BSA} can smoothly fine-tune the neural network to have the desired $\mathbf{r}_{select}$. Clearly, the regularization method of \textit{mSR} has a positive effect on simplifying how $\mathbf{r}_{select}$ should be used. 

\paragraph{Scheduled strengthening of $\lambda$:} 
The loss term during compression-friendly training is given by $\mathcal{L}(\mathbf{W}) + \lambda \sum_{l=1}^L\textit{mSR}(\mathbf{W}_l,r_l)$. As the training continues, we can expect the weight matrices to be increasingly compliant with $\mathbf{r}_{select}$, thanks to the accumulated effect of \textit{mSR} regularization. Then, a weak regularization might not be sufficient as the training continues. Comparison between fixed $\lambda$ and scheduled $\lambda$ (according to the explanation in \Cref{sec:5.1.3}) is shown in \Cref{subfig:lambda}. As expected, the scheduled strengthening of $\lambda$ is helpful for improving the compression performance.

\begin{table}[!t ]
\resizebox{\columnwidth}{!}{%
\begin{tabular}{lcccc}
\hline
\multicolumn{1}{c}{} & Method & \begin{tabular}[c]{@{}c@{}}Test\\  acc. (\%)\end{tabular} & \begin{tabular}[c]{@{}c@{}} $\Delta$ Test\\  acc. (\%)\end{tabular} & MFLOPs (rate \%) \\ \hline
\multirow{10}{*}{\rotatebox{90}{Pruning}} & Uniform & 92.80 → 89.80 & - 3.00 & 62.7 (50 \%) \\
 & LEGR \cite{chin2020towards} & 93.90 → \textbf{93.70} & - 0.20 & 58.9 (\underline{47 \%}) \\
 & ThiNet \cite{luo2017thinet} & 93.80 → 92.98 & - 0.82 & 62.7 (50 \%) \\
 & CP \cite{he2017channel} & 93.80 → 92.80 & - 1.00 & 62.7 (50 \%) \\
 & DCP \cite{zhuang2018discrimination} & 93.80 → \textbf{93.49} & - 0.31 & 62.7 (\underline{50 \%}) \\
 & AMC \cite{he2018amc} & 92.80 → 91.90 & - 0.90 & 62.7 (50 \%) \\
 & SFP \cite{he2018soft} & 93.59 → 93.35 & - 0.24 & 62.7 (50 \%) \\
 & Rethink \cite{liu2018rethinking} & 93.80 → 93.07 & - 0.73 & 62.7 (50 \%) \\
 & PFS \cite{wang2020pruning} & 93.23 → 93.05 & - 0.18 & 62.7 (50 \%) \\ 
 & CHIP \cite{sui2021chip} & 93.26 → 92.05 & - 1.21 & 34.8 (27 \%) \\ \hline 
 \multirow{4}{*}{\rotatebox{90}{Low-rank}}
 & CA \cite{alvarez2017compression} & 92.73 → 91.13 & - 1.60 & 51.4 (41 \%) \\ 
 & LC \cite{idelbayev2020low} & 92.73 → 93.10 & + 0.37 & 55.7 (44 \%) \\ 
 & $\textit{BSR}$(ours) & 92.73 → \textbf{93.53} & + 0.80 & 55.7 (\underline{44 \%}) \\ 
 & $\textit{BSR}$(ours) & 92.73 → 92.51 & - 0.22 & 32.1 (26 \%) \\ \hline
\end{tabular}}
\caption{Our method is compared with various state-of-the-art network pruning methods for ResNet56 on CIFAR-10. In the last column, “rate” stands for the percentage of the reduced FLOPs compared to the uncompressed model. A smaller rate indicates a more efficient model in terms of MFLOPs. “$\Delta$ Test acc.” stands for the difference of the test accuracy between baseline and pruned or low-rank compressed model (larger is better). For this commonly used benchmark, \textit{BSR} achieves the best performance.}
\label{tab:pruning}
\end{table}
\subsection{Low-rank compression as a highly effective tool}
Low-rank compression is only one of many ways of compressing a neural network. We first compare our results with the state-of-the-art pruning algorithms. Then, we show that our low-rank compression can be easily combined with quantization just like pruning can be. 

\subsubsection{Comparison with filtered pruning}
Both pruning and low-rank compression are techniques that are applied to weight parameters. Therefore, they cannot be used together in general. Besides, filtered pruning (structured pruning) is an actively studied topic with many known high-performance algorithms. We compared \textit{BSR} with other pruning methods including naive uniform channel number shrinkage (uniform), ThiNet \cite{luo2017thinet}, Channel Pruning (CP) \cite{he2017channel}, Discrimination-aware Channel Pruning (DCP) \cite{zhuang2018discrimination}, Soft Filter Pruning (SFP) \cite{he2018soft}, rethinking the value of network pruning (Rethink) \cite{liu2018rethinking}, Automatic Model Compression (AMC) \cite{he2018amc}, and channel independence-based pruning (CHIP)\cite{sui2021chip}. Table \ref{tab:pruning} summarizes the results.
We compared the performance drop of each method under similar FLOPs reduction rates. A smaller accuracy drop indicates a better pruning method.
When FLOPs are reduced to about 50\%, all state-of-the-art pruning algorithms exhibited performance drops, however our method ($\textit{BSR}$) showed a performance improvement (92.73 → 93.53). We also compared our method to the most recent powerful pruning algorithm (CHIP)~\cite{sui2021chip}. The CHIP algorithm can further reduce MFLOPs to 34.8 with 1.21 performance drop, and our method showed just 0.22 performance drop at similar MFLOPs (32.1MFLOPs).

\begin{table}
\resizebox{\columnwidth}{!}{%
\begin{tabular}{ccccccccc}
\hline
\multirow{2}{*}{} & \multicolumn{2}{c}{ 32bit} & \multicolumn{2}{c}{ 16bit} & \multicolumn{2}{c}{ 8bit} & \multicolumn{2}{c}{ 4bit} \\ \cline{2-9} 
 & \begin{tabular}[c]{@{}c@{}} Test\\   acc.\end{tabular}  &  \begin{tabular}[c]{@{}c@{}}  Memory\\   (Mb)\end{tabular}  & \begin{tabular}[c]{@{}c@{}}Test\\  acc.\end{tabular}  & \begin{tabular}[c]{@{}c@{}} Memory\\  (Mb)\end{tabular} &\begin{tabular}[c]{@{}c@{}}Test\\  acc.\end{tabular}  &\begin{tabular}[c]{@{}c@{}} Memory\\  (Mb)\end{tabular} & \begin{tabular}[c]{@{}c@{}}Test\\  acc.\end{tabular}  & \begin{tabular}[c]{@{}c@{}} Memory\\  (Mb)\end{tabular} \\ \hline
\begin{tabular}[c]{@{}c@{}} Base\\  network\end{tabular}& \large 0.92 & \large 3.41 & \large 0.92 & \large 1.71 & \underline{\textbf{\large 0.92}} & \underline{\textbf{\large 0.85}} & \underline{\large 0.32} & \underline{\large 0.43} \\ \hline
\begin{tabular}[c]{@{}c@{}}$\textit{BSR}$ \\ ($C_d=0.42$)\end{tabular} &  \large 0.94 & \large 1.98 & \large 0.94 & \large 0.99 & \textbf{\large 0.94} & \textbf{\large 0.49} & \large 0.32 & \large 0.25 \\
\begin{tabular}[c]{@{}c@{}}$\textit{BSR}$\\ ($C_d=0.56$)\end{tabular} & \large 0.94 & \large 1.50 & \large 0.94 & \large 0.75 & \textbf{\large 0.93} & \textbf{\large 0.38} & \large 0.32 & \large 0.19 \\
\begin{tabular}[c]{@{}c@{}}$\textit{BSR}$ \\ ($C_d=0.63$)\end{tabular} & \large 0.94 & \large 1.27 & \large 0.94 & \large 0.64 & \textbf{\large 0.93} & \textbf{\large 0.32} & \large 0.31 & \large 0.16 \\
\begin{tabular}[c]{@{}c@{}}$\textit{BSR}$ \\ ($C_d=0.77$)\end{tabular} & \large 0.93 & \large 0.79 & \large 0.93 & \large 0.40 & \underline{\textbf{\large 0.92}} & \underline{\textbf{\large 0.20}} & \large 0.33 & \large 0.10 \\
\begin{tabular}[c]{@{}c@{}}$\textit{BSR}$ \\ ($C_d=0.84$)\end{tabular} & \large 0.91 & \large 0.54 & \large 0.91 & \large 0.27 & \textbf{\large 0.90} & \textbf{\large 0.14} & \large 0.32 & \large 0.07 \\
\begin{tabular}[c]{@{}c@{}}$\textit{BSR}$ \\ ($C_d=0.90$)\end{tabular} & \large 0.89 & \large 0.35 & \large 0.89 & \large 0.17 & \textbf{\large 0.83} & \textbf{\large 0.09} & \large 0.31 & \large 0.04 \\ \hline
\multicolumn{1}{l}{} & \multicolumn{1}{l}{} & \multicolumn{1}{l}{} & \multicolumn{1}{l}{} & \multicolumn{1}{l}{} & \multicolumn{1}{l}{} & \multicolumn{1}{l}{} & \multicolumn{1}{l}{} & \multicolumn{1}{l}{}
\end{tabular}}
\caption{Quantization results of ResNet56 on CIFAR-10. The test accuracy and memory size are represented as bit-width changes.}
\label{tab:quantization}
\end{table}

\subsubsection{Combined use with quantization}
As known well, low-rank compression can be used together with quantization. The performance for using both together is shown in \Cref{tab:quantization}. Even without a performance loss, we can observe that memory usage can be additionally reduced to one-fifth compared to $\textit{BSR}$ only. In particular, when comparing the case with the same accuracy (when $C_{d}$ = 0.77) as the base network, the memory usage was reduced to 1/17 (0.2 Mb) compared to the base network (3.41 Mb). We were also able to reach a range of memory usage that could not be attained by quantization alone by using \textit{BSR} together. Note that \textit{BSR} allows any compression ratio while quantization does not. This feature can make using deep learning models on the edge devices much easier. Quantization and our low-rank compression are extremely easy to use and can be applied for practically any situation. 

\subsection{Limitations and future works}
It might be possible to improve \textit{BSA} in a few different ways. While our modified stable rank works well, it might be possible to identify a rank surrogate with a better learning dynamics. While we choose only one $\mathbf{r}_{selected}$ and perform only a single compression-friendly training, it might be helpful to choose multiple rank vectors, train all, and choose the best. This can be an obvious way of improving performance at the cost of an extra computation. Pruning and low-rank compression cannot be used at the same time, but it might be helpful to apply them sequentially. In general, combining multiple compression techniques to generate a synergy remains as a future work.

%% file: 7_conclusion.tex
\section{Conclusion}
\label{sec:conclusion}
We have introduced a new low-rank compression method called \textit{BSR}. Its main improvements compared to the previous works are in the rank selection algorithm and the rank regularized training. The design of modified beam-search is based on the idea that beam-search is a superior way of balancing search performance and search speed. Modifications such as the introduction of level step size $s$ and the compression rate constraint play important roles.
The design of modified stable rank is based on a careful analysis on how weight matrix's rank is regularized. With our best knowledge, our modified stable rank is the first regularization method that truly controls the rank. As the result, \textit{BSR} performs very well.

%% file: 8_appendix.tex
\newpage
\onecolumn
\begin{center}
\textbf{\Large Supplementary materials for the paper \\ ```A Highly Effective Low-Rank Compression of Deep Neural Networks \\
    with Modified Beam-Search and Modified Stable Rank''}
\end{center}
\appendix

\section{Derivative of mSR}
The gradient of $\textit{mSR}$ is derived by first decomposing $\mathbf{W}_l = \mathbf{U}_l\mathbf{\Sigma }_l\mathbf{V}_l^{T}$ into two parts by allocating the first $r_l$ dimensions into $\mathbf{W}_l^{1:r_l}$ and the remaining dimensions into $\mathbf{W}_l^{r_l:R_l}$ as below.
\begin{align*}
\small
\begin{split}
    \mathbf{W}_l &= \mathbf{W}_l^{1:r_l} \mathbf{W}_l^{r_l:R_l} \\
            &  = \mathbf{U}_l^{1:r_l} \mathbf{\Sigma}_{l}^{1:r_l} (\mathbf{V}_l^{1:r_l})^{T} + \mathbf{U}_l^{r_l:R_l} \mathbf{\Sigma}_{l}^{r_l:R_l} (\mathbf{V}_l^{r_l:R_l})^{T}
\end{split}
\end{align*}
Then, the gradient can be derived as the following.
\begin{align*}
    \begin{split}
        & \qquad \frac{\partial \textit{mSR}(\mathbf{W}_l,r_l)}{\partial \mathbf{W}_l} = \frac{\partial \left(\frac{tr(\mathbf{\Sigma}_l^{r_l:R_l})}{tr(\mathbf{\Sigma}_l^{1:r_l})}\right)}{\partial \mathbf{W}_l} \\
        & = \frac{1}{(tr(\mathbf{\Sigma}_l^{1:r_l}))^{2}}\left(tr(\mathbf{\Sigma}_l^{1:r_l})\frac{\partial tr(\mathbf{\Sigma}_l^{r_l:R_l})}{\partial \mathbf{W}_l}-tr(\mathbf{\Sigma}_l^{r_l:R_l})\frac{\partial tr(\mathbf{\Sigma}_l^{1:r_l})}{\partial \mathbf{W}_l}\right) \\
        & = \frac{tr(\mathbf{\Sigma}_l^{r_l:R_l})}{tr(\mathbf{\Sigma}_l^{1:r_l})}\biggl[\frac{1}{tr(\mathbf{\Sigma}_l^{r_l:R_l})}\biggl(\mathbf{V_{l}^{r_l:R_l}}(\mathbf{U}_{l}^{r_l:R_l})^{T} - \frac{\partial tr((\mathbf{U}_{l}^{r_l:R_l})^{T} \mathbf{W}_{l}^{r_l:R_l}\mathbf{V}_{l}^{r_l:R_l})}{\partial \mathbf{W}_l}\biggr) \\ 
        & \qquad - \frac{1}{tr(\mathbf{\Sigma}_{l}^{1:r_l})}\biggl(\mathbf{V}_{l}^{1:r_l}(\mathbf{U}_{l}^{1:r_l})^{T}-\frac{\partial tr((\mathbf{U}_{l}^{1:r_l})^{T}\mathbf{W}_{l}^{r_l:R_l}\mathbf{V}_{l}^{1:r_l})}{\partial \mathbf{W}_l}\biggr)\biggr] \\
        & = \frac{tr(\mathbf{\Sigma}_{l}^{r_l:R_l})}{tr(\mathbf{\Sigma}_{l}^{1:r_l})}\left(\frac{\mathbf{U}_{l}^{r_l:R_l}(\mathbf{V}_{l}^{r_l:R_l})^{T}}{tr(\mathbf{\Sigma}_{l}^{r_l:R_l})} - \frac{\mathbf{U}_{l}^{1:r_l}(\mathbf{V}_{l}^{1:r_l})^{T}}{tr(\mathbf{\Sigma}_{l}^{1:r_l})}\right). 
    \end{split}
\end{align*}

\section{Experimental results for FLOPs}

\subsection{Number of floating point operations~(FLOPs) computation}
In the literature, there is no clear consensus on how to compute the total number of floating point operations~(FLOPs) in the forward pass of neural network. While some authors define this number as the total number of multiplications and additions~\cite{ye2018rethinking}, others assume that multiplications and additions can be fused and compute one multiplication and addition as a single operation~\cite{he2016deep}. In this work, we use the second definition of FLOPs.

\subsubsection{FLOPs in a fully-connected layer}
For a fully-connected layer with weight $\mathbf{W}\in\mathbb{R}^{m\times n}$ and bias $\mathbf{b}\in\mathbb{R}^{m}$, FLOPs can be calculated as below. 
\begin{equation*}
    \textit{fc-FLOPs}(\mathbf{W}, \mathbf{b})= m \times(n-1)+m = mn
\end{equation*}

\subsubsection{FLOPs in a convolutional layer}
For a convolutional layer with parameters $\mathbf{W}$ and $\mathbf{b}$, linear mapping is applied $M$ times. Thus, FLOPs can be calculated as below. 
\begin{equation*}
    \textit{conv-FLOPs}=\textit{fc-FLOPs}(\mathbf{W},\mathbf{0})\times M + \textit{fc-FLOPs}(\mathbf{0},\mathbf{b})
\end{equation*}
We exclude the batch-normalization~(BN) and concatenation/copy-operation. For BN layers, the BN parameters can be fused into the weights and biases of the preceding layer, and therefore no special treatment is required. For concatenation, copy-operations, and non-linearities, we assume zero FLOPs because of its negligible cost.

\subsection{MNIST}
\begin{figure}[!t]
    \centering
    \includegraphics[width=0.30\textwidth]{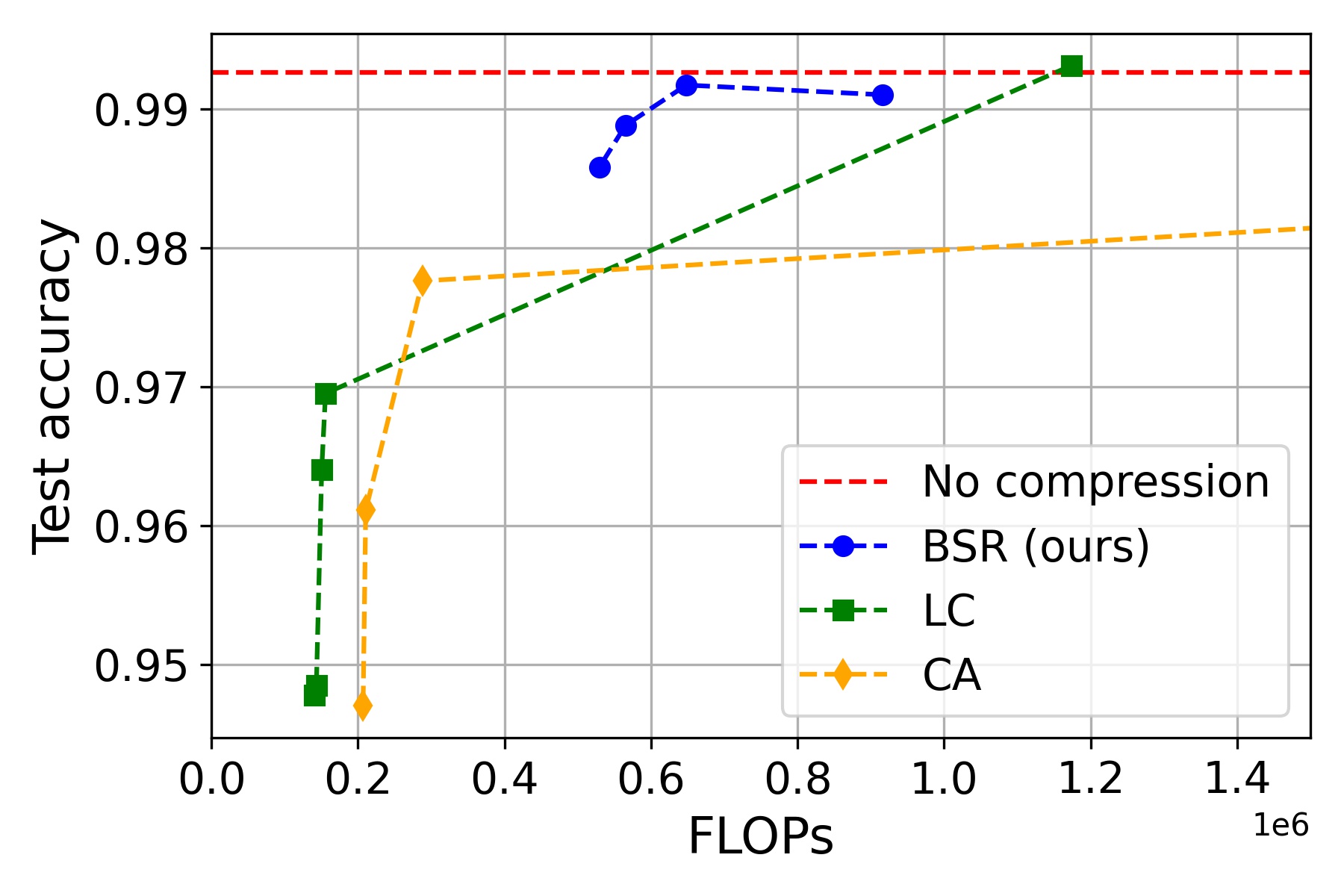}
    \caption{Comparison of $\textit{BSR}$ with CA and LC for LeNet5 on MNIST.}
    \label{fig:LeNet-5 on MNIST_flops}
\end{figure}

The results for LeNet5 on MNIST are shown in \Cref{fig:LeNet-5 on MNIST_flops}. By considering FLOPs instead of the compression ratio, the resulting range of each approach varied. Nonetheless, it is not difficult to confirm that both of $\textit{BSR}$ and LC outperform CA. Between BSR and LC, BSR outperforms LC for the entire $\textit{BSR}$ evaluation range.

\subsection{CIFAR-10 and CIFAR-100}
\begin{figure*}
     \centering
     \begin{subfigure}[b]{0.3\textwidth}
         \centering
         \includegraphics[width=\textwidth]{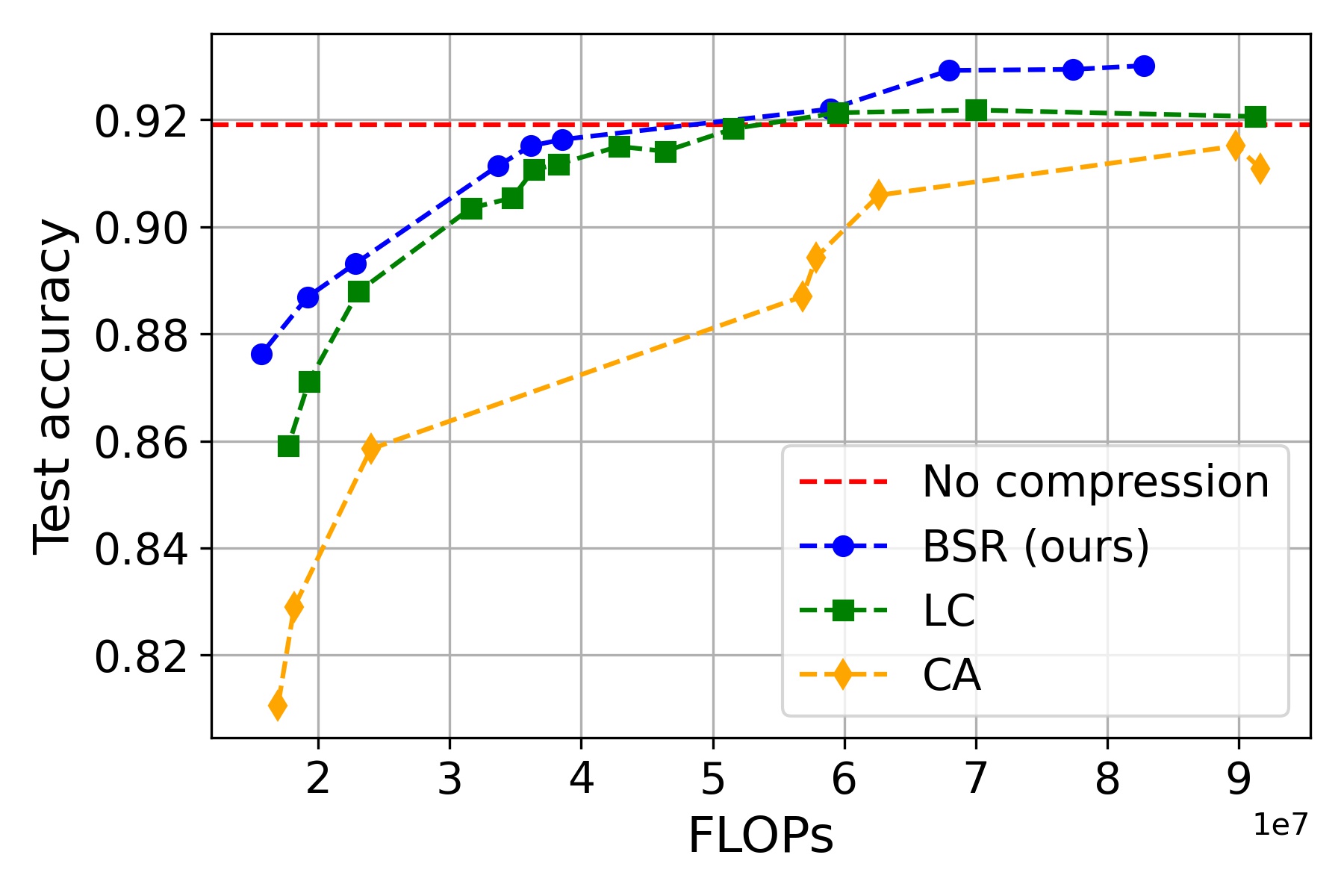}
         \caption{ResNet32 on CIFAR-10}
         \label{fig:ResNet32 on CIFAR10 for flops}
     \end{subfigure}
     \hfill
     \begin{subfigure}[b]{0.3\textwidth}
         \centering
         \includegraphics[width=\textwidth]{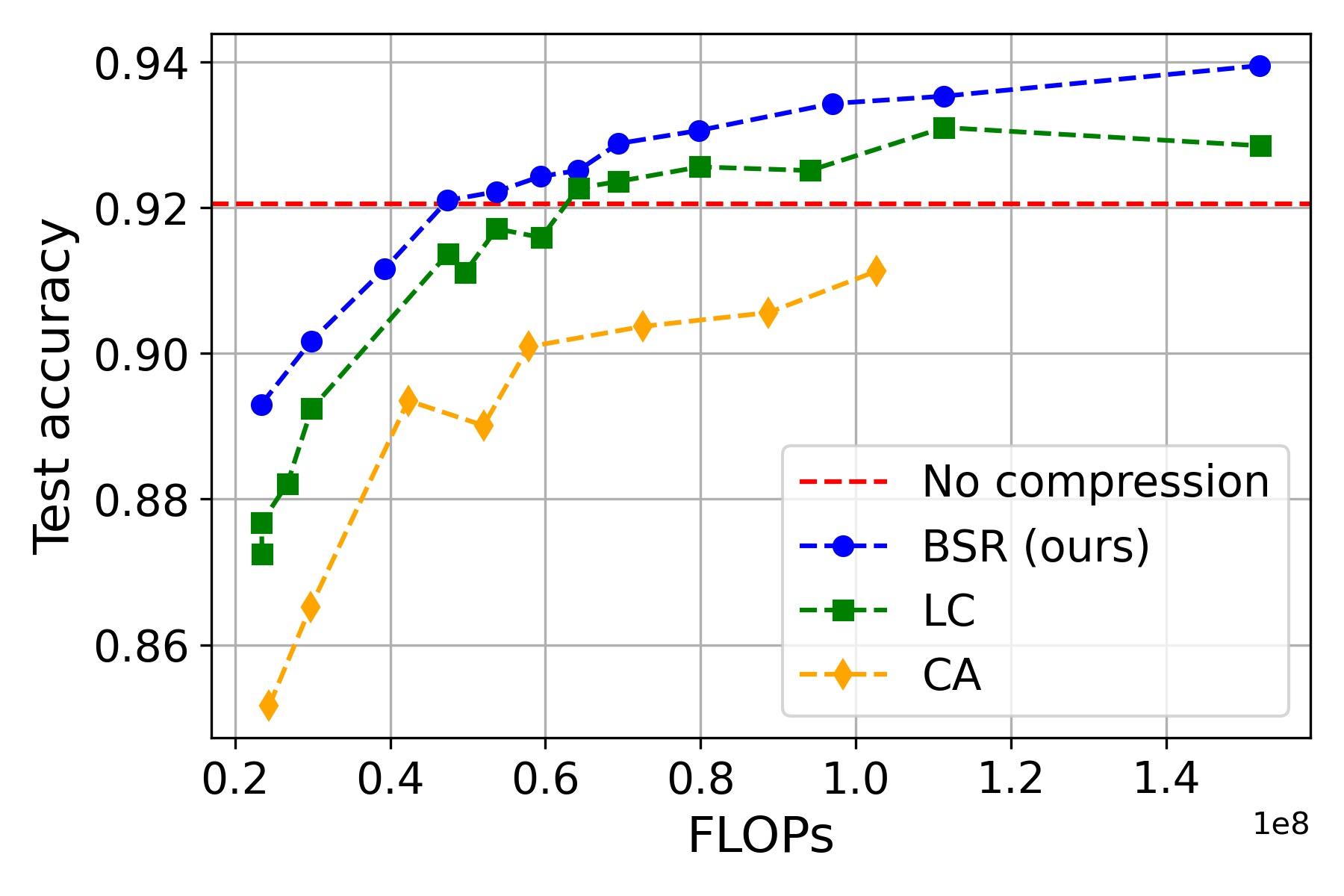}
         \caption{ResNet56 on CIFAR-10}
         \label{fig:ResNet56 on CIFAR10 for flops}
     \end{subfigure}
      \hfill
     \begin{subfigure}[b]{0.3\textwidth}
         \centering
         \includegraphics[width=\textwidth]{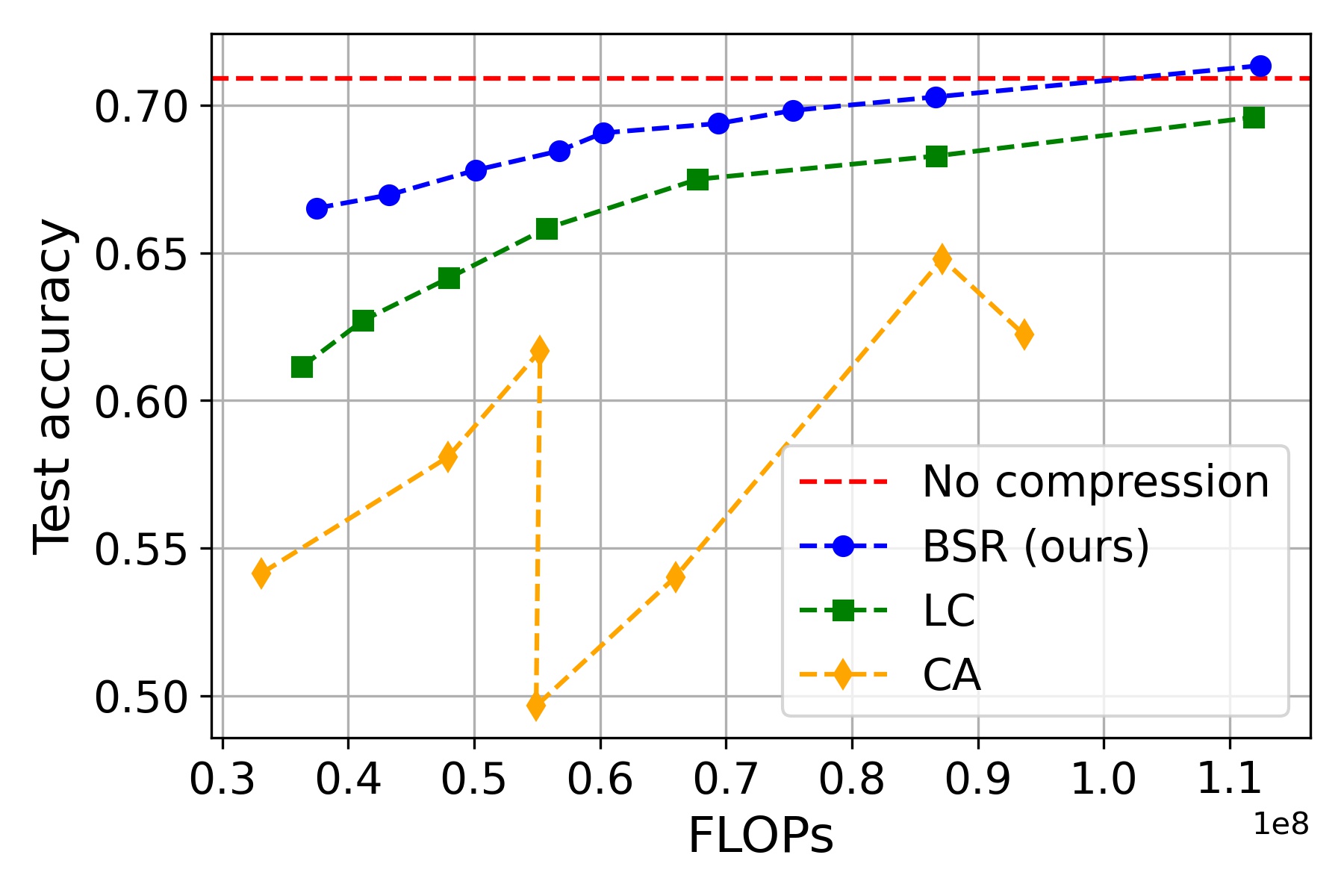}
         \caption{ResNet56 on CIFAR-100}
         \label{fig:ResNet56 on CIFAR100 for flops}
     \end{subfigure}
\caption{Comparison of $\textit{BSR}$ with CA and LC for (a) ResNet32 on CIFAR-10, (b) ResNet56 on CIFAR-10, and (c) ResNet56 on CIFAR-100.}
\label{fig:ResNet on CIFAR for flops}
\end{figure*}

The results for ResNet32/ResNet56 on CIFAR-10 and the results for ResNet56 on CIFAR-100 are shown in \Cref{fig:ResNet on CIFAR for flops}. For ResNet, \textit{BSR} performs a low-rank compression over the convolutional layers. While LC directly considered FLOPs when selecting rank, we selected rank by considering only the validation accuracy. Nonetheless, we achieve higher test accuracies.
$\textit{BSR}$ can even improve the no-compression accuracy when FLOPs is not too small. For instance, test accuracy in \Cref{fig:ResNet56 on CIFAR10 for flops} is improved from 0.92 to 0.94 by \textit{BSR} when the FLOPs is around 140~MFLOPs. LC is also able to improve the accuracy, but the gain is smaller than \textit{BSR}. This implies that the regularized training methods can have a positive influence on learning dynamics. Similarly, in \Cref{fig:ResNet32 on CIFAR10 for flops} and \Cref{fig:ResNet56 on CIFAR100 for flops}, the test accuracy is improved over no-compression when the flops is not too small. 

\subsection{ImageNet}
\begin{figure}[!t]
    \centering
    \includegraphics[width=0.30\textwidth]{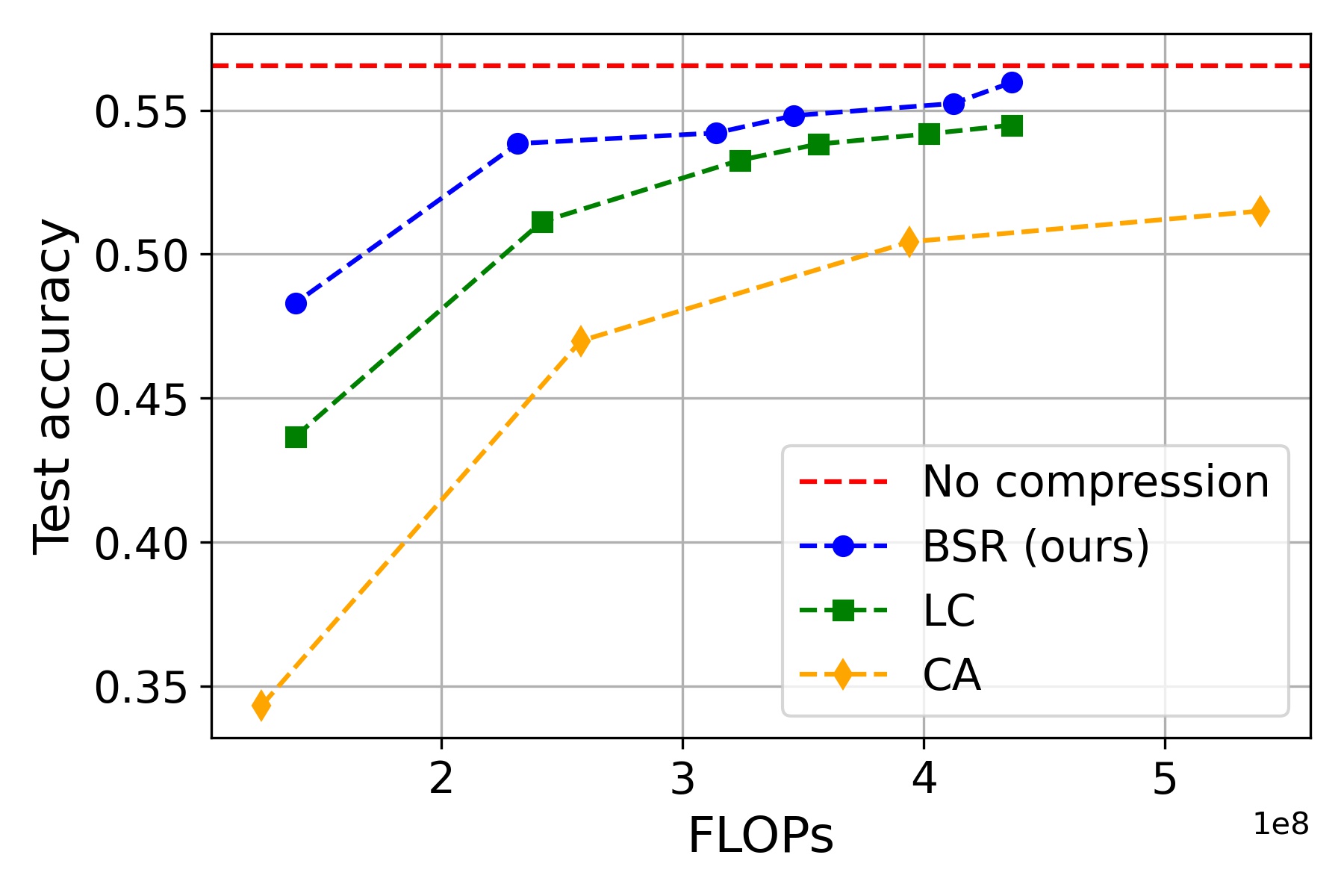}
    \caption{Comparison of $\textit{BSR}$ with CA and LC for AlexNet on ImageNet (Top-1 performance).}
    \label{fig:Alexnet on ImageNet for flops}
\end{figure}

The results for AlexNet on ImageNet are shown in \Cref{fig:Alexnet on ImageNet for flops}. 
As in the compression ratio vs test accuracy, we used the pre-trained Pytorch AlexNet network as the base network. 

ImageNet is a much more complex dataset than MNIST or CIFAR.
The performance curves show similar patterns as in \Cref{fig:ResNet on CIFAR for flops}, confirming that \textit{BSR} works well for complex datasets.

\label{sec:imagenet_qt}
\section{Quantization results of AlexNet on ImageNet}
In this section, we combine low-rank compression and quantization for ImageNet. 
The performance for using both is shown in \Cref{tab:quantization_alexnet}. 
Although there can be a slight performance loss, we can observe a reduction in memory usage by using quantization in addition to BSR.
When $C_d=0.56$, we can save more than half of the memory by reducing quantization from 32bits to 16bits.
Similar to the result of ResNet56, we were also able to reach a range of memory usage that could not be attained by quantization alone by using \textit{BSR} together. 
Even with more realistic datasets and larger networks, using BSR and quantization together is effective in reducing memory usage.
This feature can make it much easier to use deep learning models on the edge devices. 
Quantization and our low-rank compression are extremely easy to use and can be applied for practically any situation.

\begin{table}
\centering
    \resizebox{0.7\columnwidth}{!}{%
    \begin{tabular}{ccccccccc}
        \hline
        \multirow{2}{*}{} & \multicolumn{2}{c}{32 bits} & \multicolumn{2}{c}{16 bits} & \multicolumn{2}{c}{8 bits} & \multicolumn{2}{c}{4 bits} \\ \cline{2-9} 
         & \begin{tabular}[c]{@{}c@{}} Test\\   acc.\end{tabular}  &  \begin{tabular}[c]{@{}c@{}}  Memory\\   (Mb)\end{tabular}  & \begin{tabular}[c]{@{}c@{}}Test\\  acc.\end{tabular}  & \begin{tabular}[c]{@{}c@{}} Memory\\  (Mb)\end{tabular} &\begin{tabular}[c]{@{}c@{}}Test\\  acc.\end{tabular}  &\begin{tabular}[c]{@{}c@{}} Memory\\  (Mb)\end{tabular} & \begin{tabular}[c]{@{}c@{}}Test\\  acc.\end{tabular}  & \begin{tabular}[c]{@{}c@{}} Memory\\  (Mb)\end{tabular} \\ \hline
        \begin{tabular}[c]{@{}c@{}} Base\\  network\end{tabular}& \large 0.56 & \large 244.40 & \large 0.56 & \large 122.20 & \underline{\textbf{\large 0.54}} & \underline{\textbf{\large 61.10}} & \underline{\large 0.04} & \underline{\large 30.55} \\ \hline
        \begin{tabular}[c]{@{}c@{}}$\textit{BSR}$ \\ ($C_d=0.56$)\end{tabular} &  \large 0.55 & \large 108.17 & \large 0.55 & \large 54.09 & \textbf{\large 0.52} & \textbf{\large 27.04} & \large 0.03 & \large 13.52 \\
        \begin{tabular}[c]{@{}c@{}}$\textit{BSR}$\\ ($C_d=0.63$)\end{tabular} & \large 0.55 & \large 91.46 & \large 0.54 & \large 45.72 & \textbf{\large 0.52} & \textbf{\large 22.86} & \large 0.03 & \large 11.43 \\
        \begin{tabular}[c]{@{}c@{}}$\textit{BSR}$ \\ ($C_d=0.71$)\end{tabular} & \large 0.55 & \large 71.50 & \large 0.55 & \large 35.75 & \textbf{\large 0.51} & \textbf{\large 17.88} & \large 0.03 & \large 8.94 \\
        \begin{tabular}[c]{@{}c@{}}$\textit{BSR}$ \\ ($C_d=0.75$)\end{tabular} & \large 0.54 & \large 60.34 & \large 0.54 & \large 30.17 & \underline{\textbf{\large 0.51}} & \underline{\textbf{\large 15.09}} & \large 0.03 & \large 7.54 \\
        \begin{tabular}[c]{@{}c@{}}$\textit{BSR}$ \\ ($C_d=0.84$)\end{tabular} & \large 0.54 & \large 38.38 & \large 0.52 & \large 19.19 & \textbf{\large 0.48} & \textbf{\large 9.59} & \large 0.03 & \large 4.80 \\
        \begin{tabular}[c]{@{}c@{}}$\textit{BSR}$ \\ ($C_d=0.91$)\end{tabular} & \large 0.48 & \large 22.06 & \large 0.45 & \large 11.03 & \textbf{\large 0.43} & \textbf{\large 5.52} & \large 0.02 & \large 2.76 \\ \hline
        \multicolumn{1}{l}{} & \multicolumn{1}{l}{} & \multicolumn{1}{l}{} & \multicolumn{1}{l}{} & \multicolumn{1}{l}{} & \multicolumn{1}{l}{} & \multicolumn{1}{l}{} & \multicolumn{1}{l}{} & \multicolumn{1}{l}{}
    \end{tabular}}
    \caption{Quantization results of AlexNet on ImageNet. The test accuracy and memory size are represented as bit-width changes.}
    \label{tab:quantization_alexnet}
\end{table}

\section{Additional ablation tests}
Additional results are provided for adjusting $K$ and $s$.
\subsection{The effect of level step size $s$}
\begin{figure}[!ht]
    \centering
    \begin{subfigure}{0.3\textwidth}
        \centering
        \includegraphics[width=\textwidth]{Figures/BSR_k_1.jpg}
        \caption{$K=1$} 
        \label{subfig:k=1_app}
    \end{subfigure}
    \hfill
    \begin{subfigure}{0.3\textwidth}
        \centering
        \includegraphics[width=\textwidth]{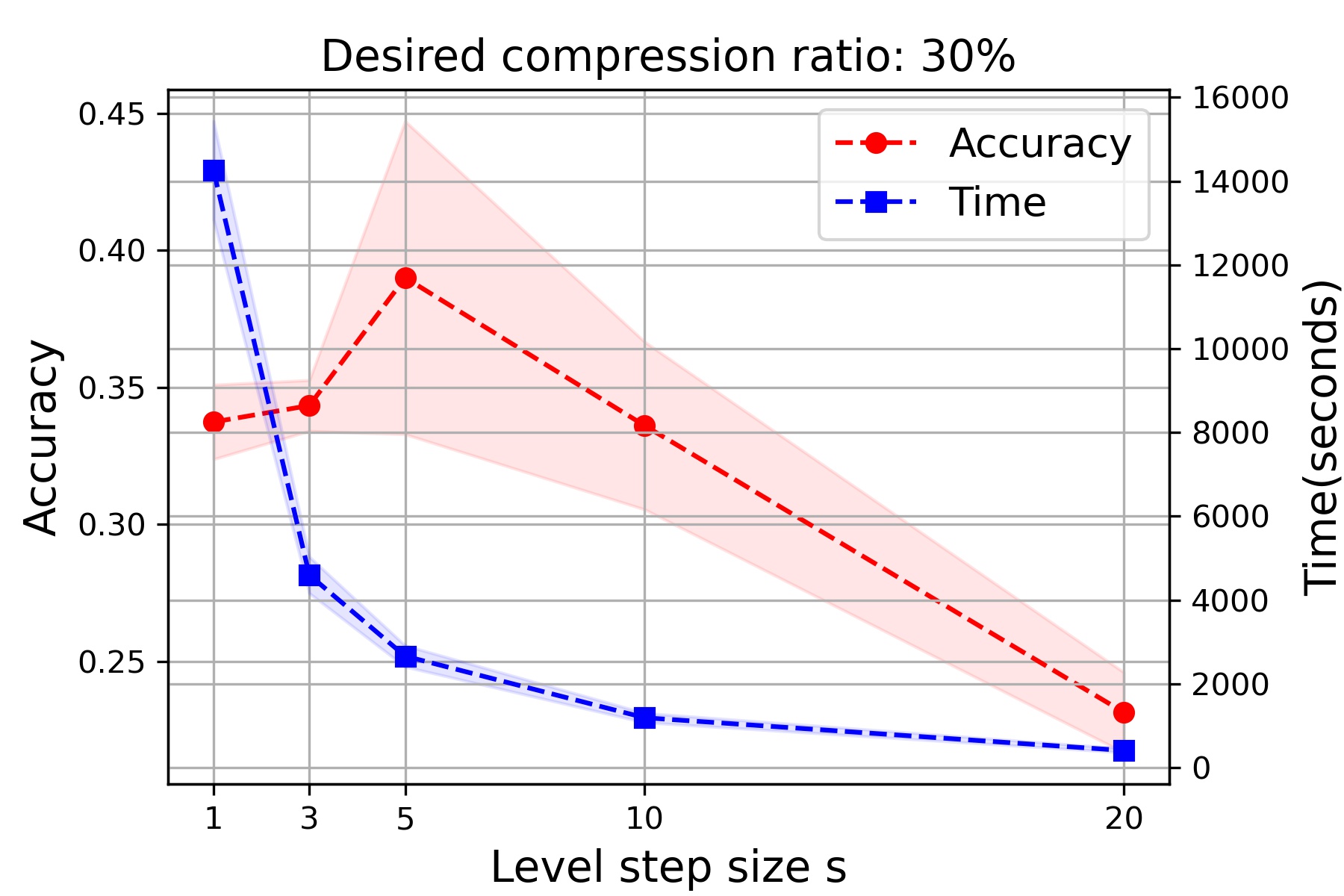}
        \caption{$K=3$}
        \label{subfig:k=3}
    \end{subfigure}
    \hfill
    \begin{subfigure}{0.3\textwidth}
        \centering
        \includegraphics[width=\textwidth]{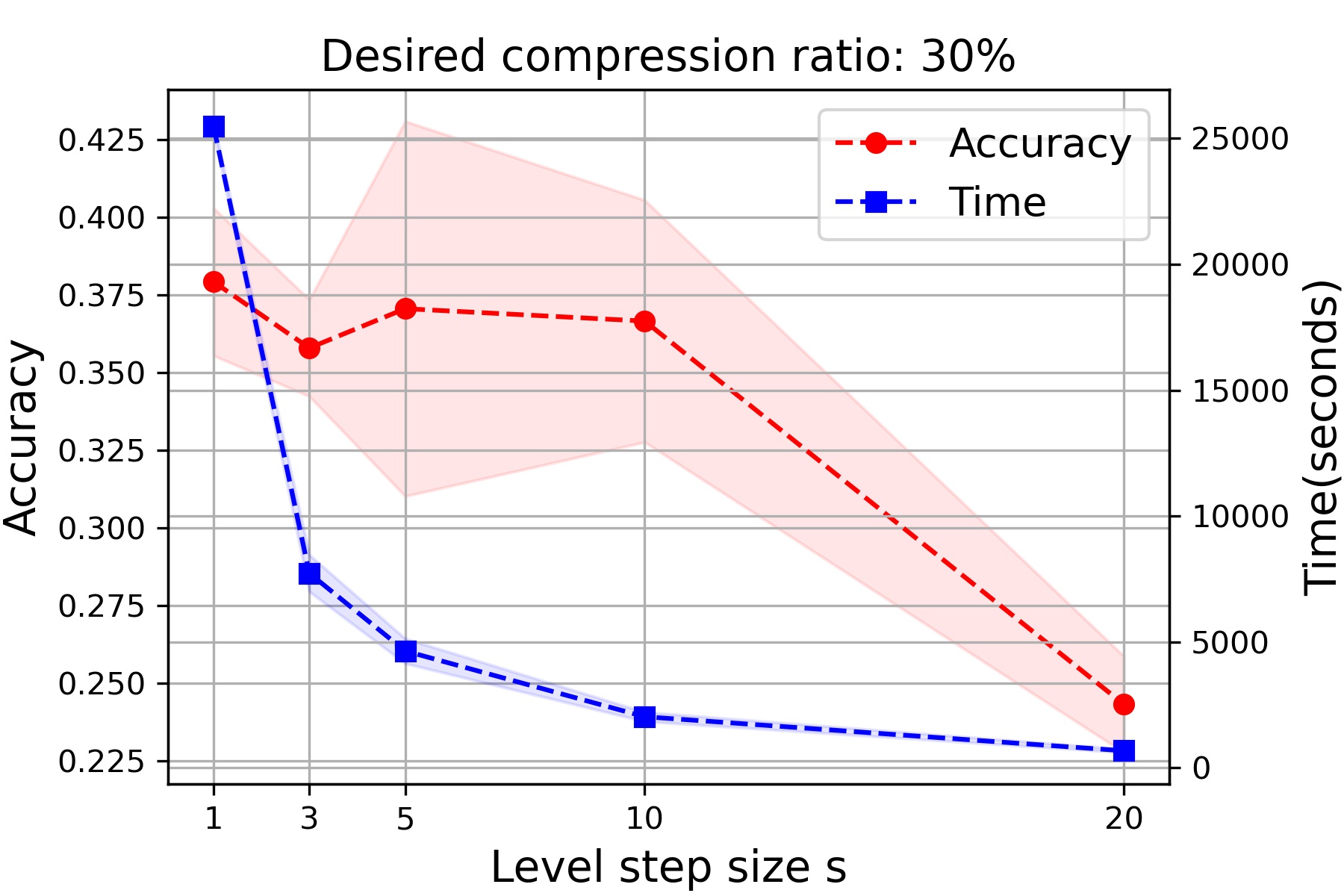}
        \caption{$K=5$}
        \label{subfig:k=5}
    \end{subfigure}
    \hfill
    \begin{subfigure}{0.3\textwidth}
        \centering
        \includegraphics[width=\textwidth]{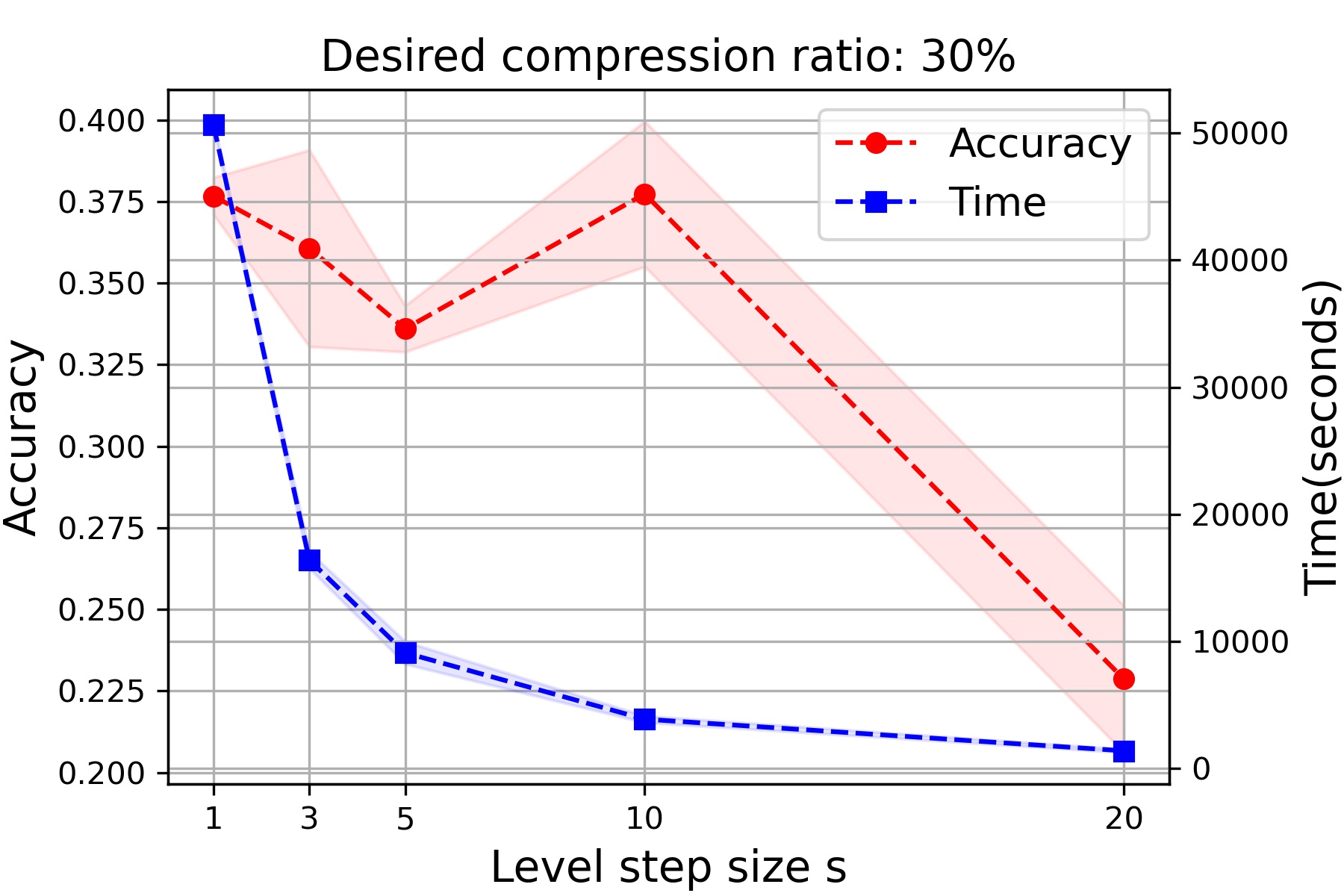}
        \caption{$K=10$}
        \label{subfig:k=10}
    \end{subfigure}
    \hfill
    \begin{subfigure}{0.3\textwidth}
        \centering
        \includegraphics[width=\textwidth]{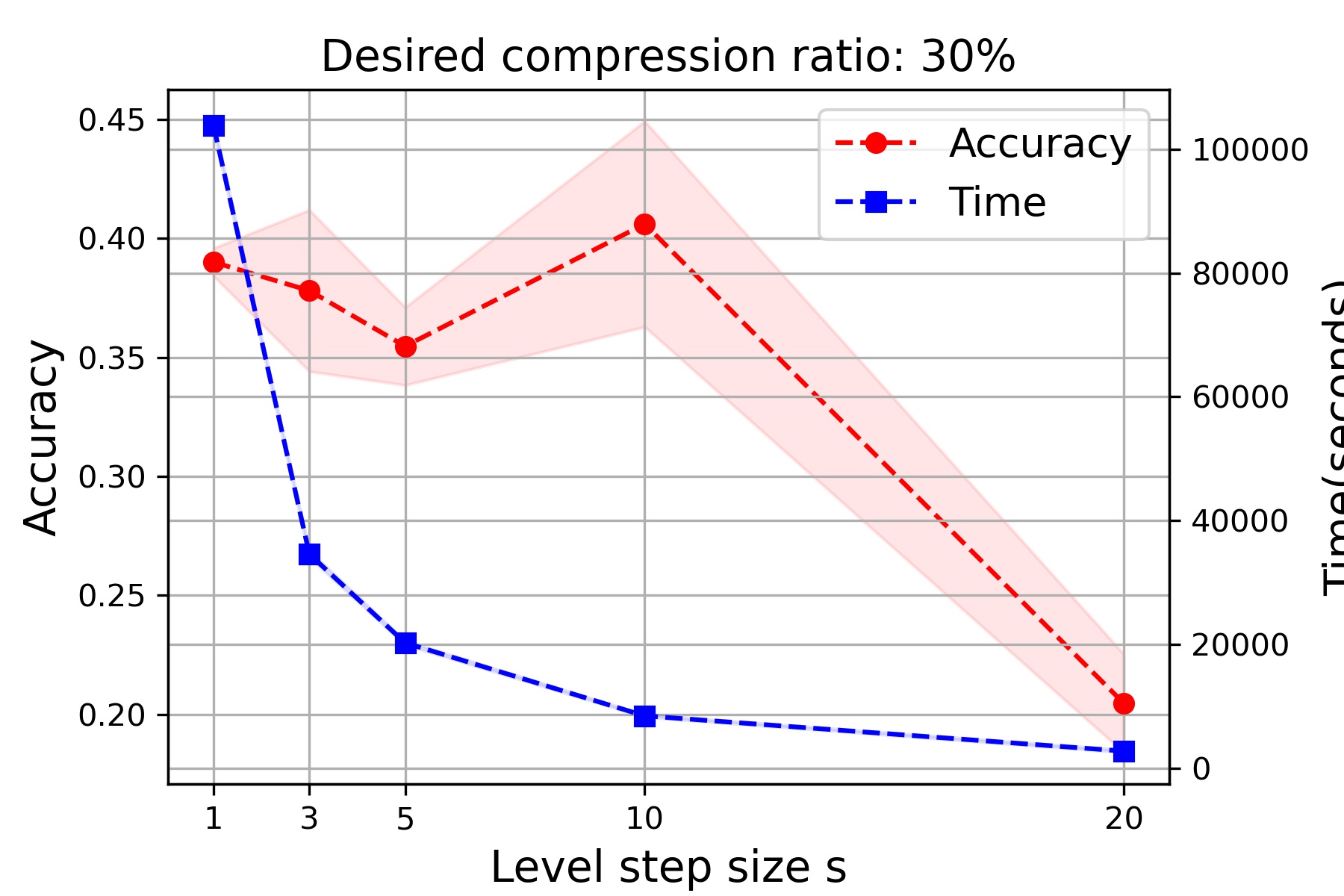}
        \caption{$K=20$}
        \label{subfig:k=20}
    \end{subfigure}
\caption{The effect of $s$ parameters on $\textit{mBS}$'s performance for a fixed $K$: a base neural network (ResNet56 on CIFAR-100) was truncated by the selected ranks and no further fine-tuning was applied for this analysis. Performance and search speed change as a function of $s$: (a) $K$ is fixed at 1, (b) $K$ is fixed at 3, (c) $K$ is fixed at 5, (d) $K$ is fixed at 10, (e) $K$ is fixed at 20.}
\label{fig:fixed K}
\end{figure}

We checked the effect of the level step size $s$ for a fixed beam size $K$. The analysis was repeated for a range of $K$, and the results are shown in \Cref{fig:fixed K}. From the results, it can be confirmed that it is desirable to choose an $s$ that is not too large.

\begin{figure}[!ht]
    \centering
    \begin{subfigure}{0.3\textwidth}
        \centering
        \includegraphics[width=\textwidth]{Figures/BSR_step_1.jpg}
        \caption{$s=1$} 
        \label{subfig:s=1}
    \end{subfigure}
    \hfill
    \begin{subfigure}{0.3\textwidth}
        \centering
        \includegraphics[width=\textwidth]{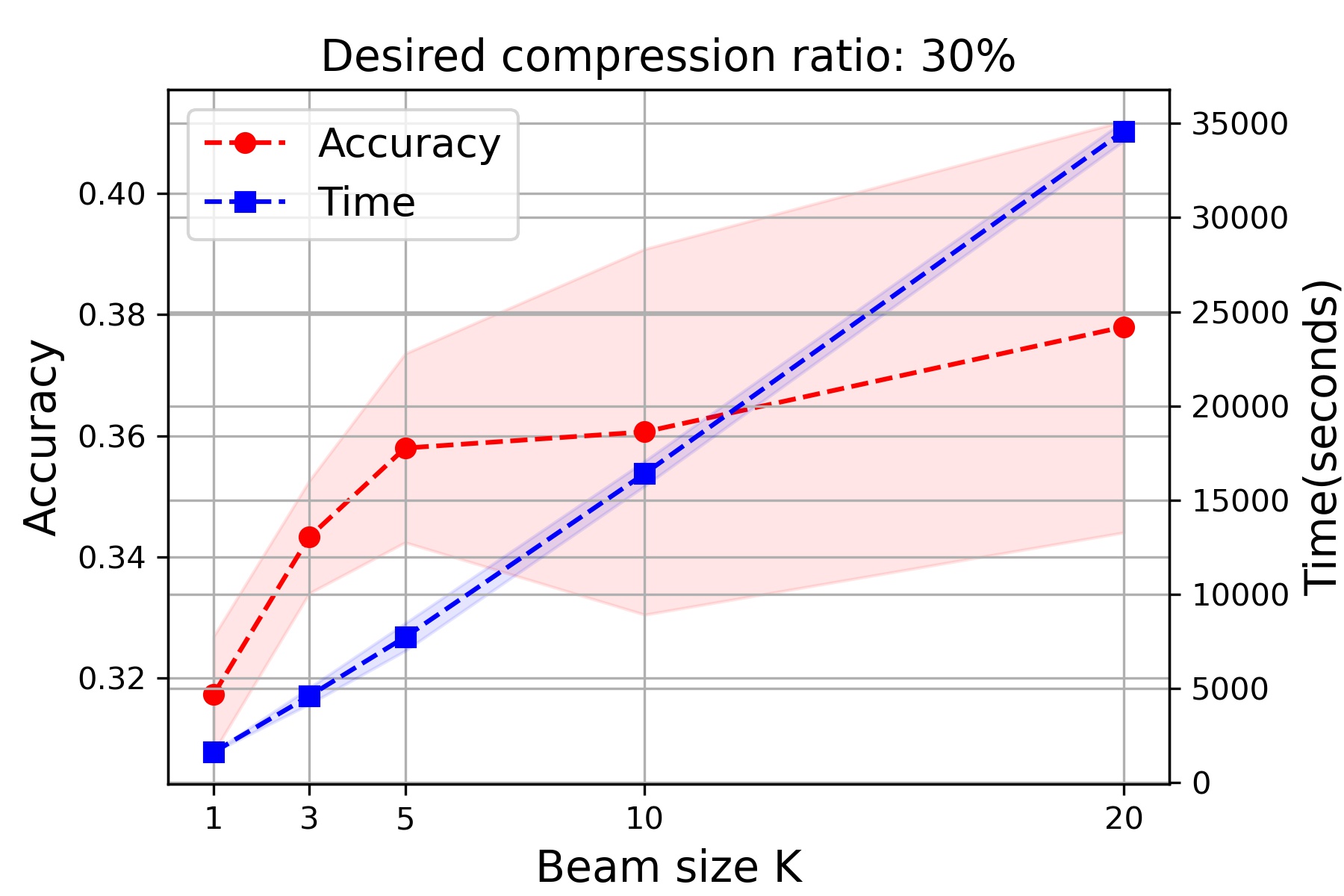}
        \caption{$s=3$}
        \label{subfig:s=3}
    \end{subfigure}
    \hfill
    \begin{subfigure}{0.3\textwidth}
        \centering
        \includegraphics[width=\textwidth]{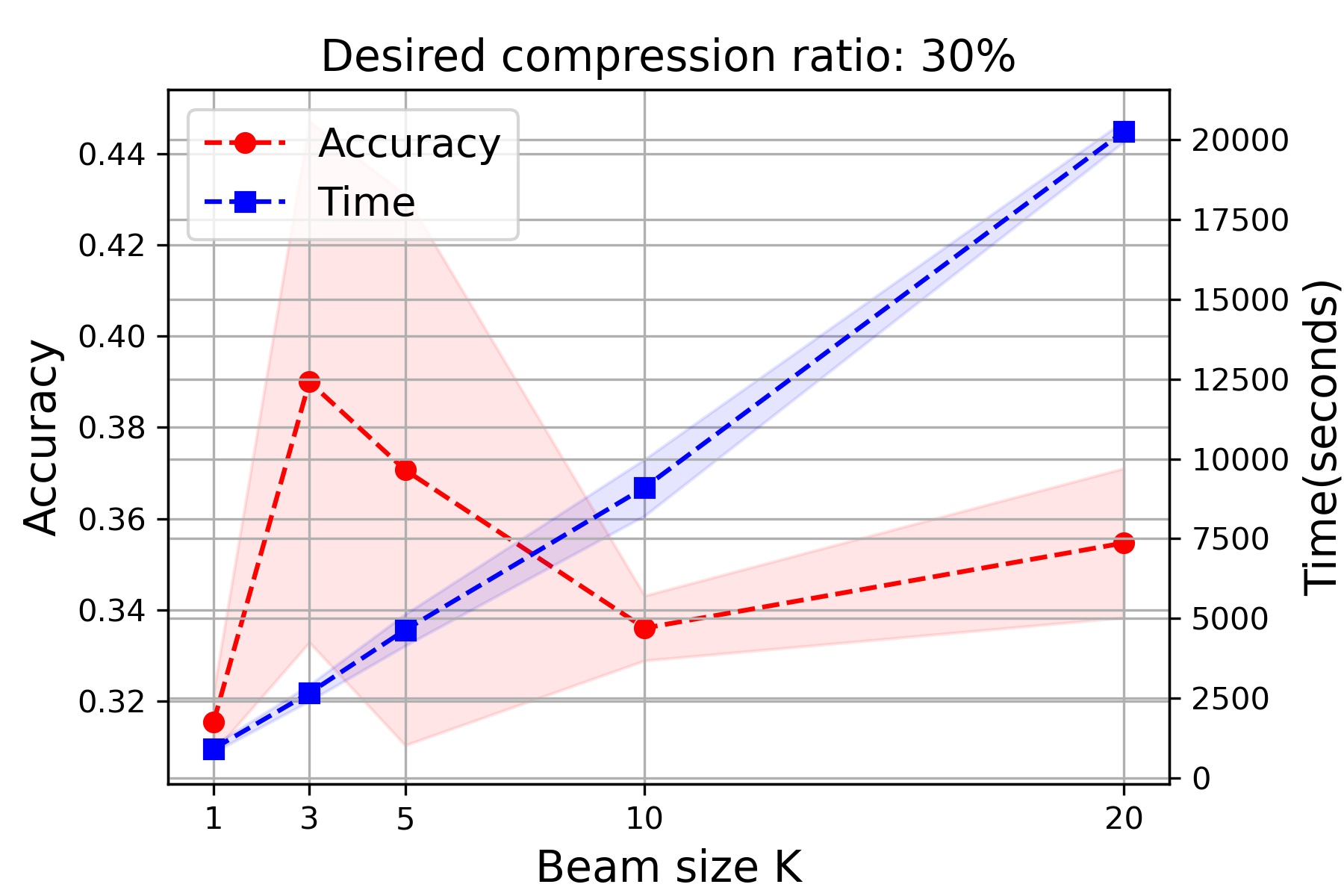}
        \caption{$s=5$}
        \label{subfig:s=5}
    \end{subfigure}
    \hfill
    \begin{subfigure}{0.3\textwidth}
 
        \includegraphics[width=\textwidth]{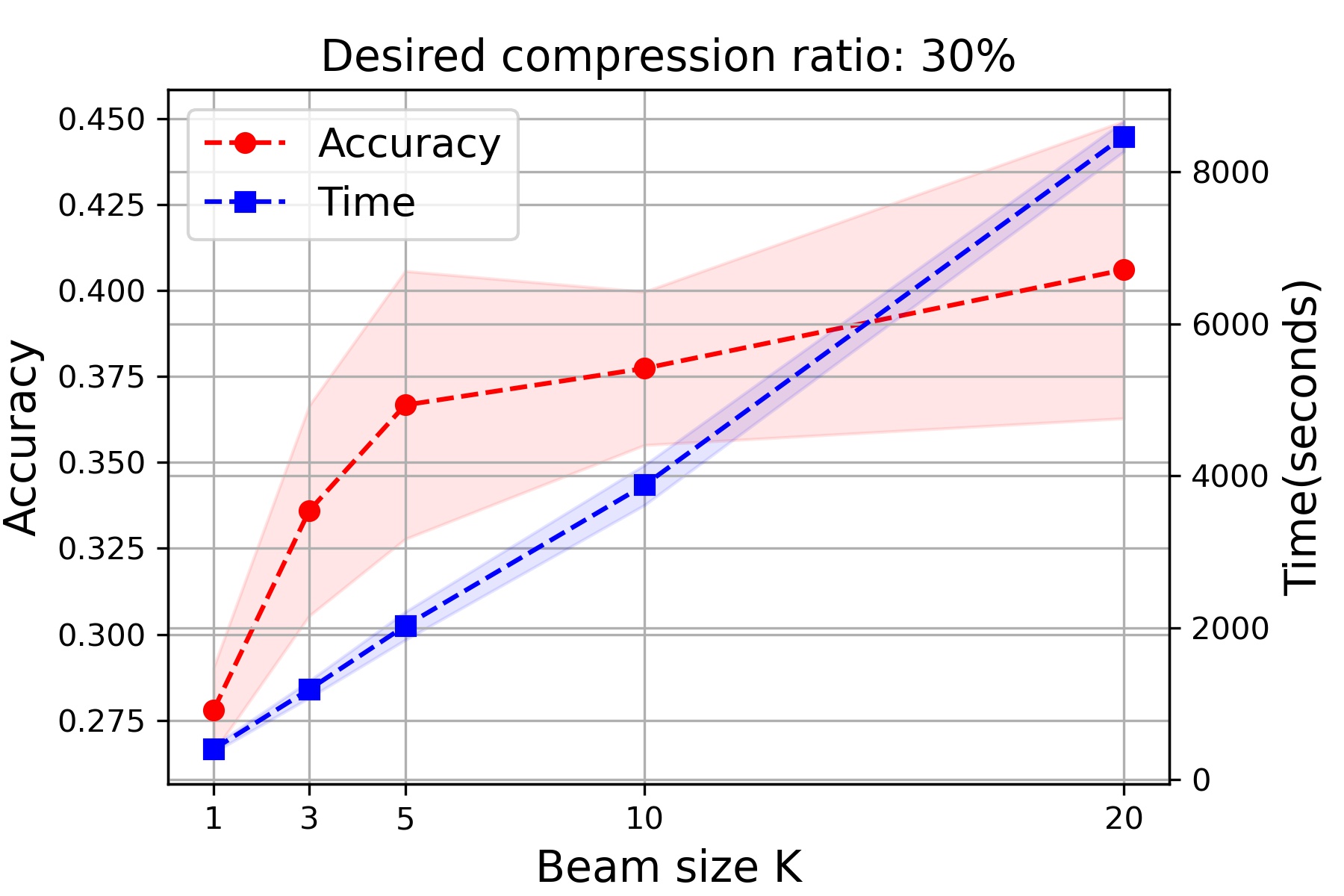}
        \caption{$s=10$}
        \label{subfig:s=10}
    \end{subfigure}
    \hfill
    \begin{subfigure}{0.3\textwidth}
    
        \includegraphics[width=\textwidth]{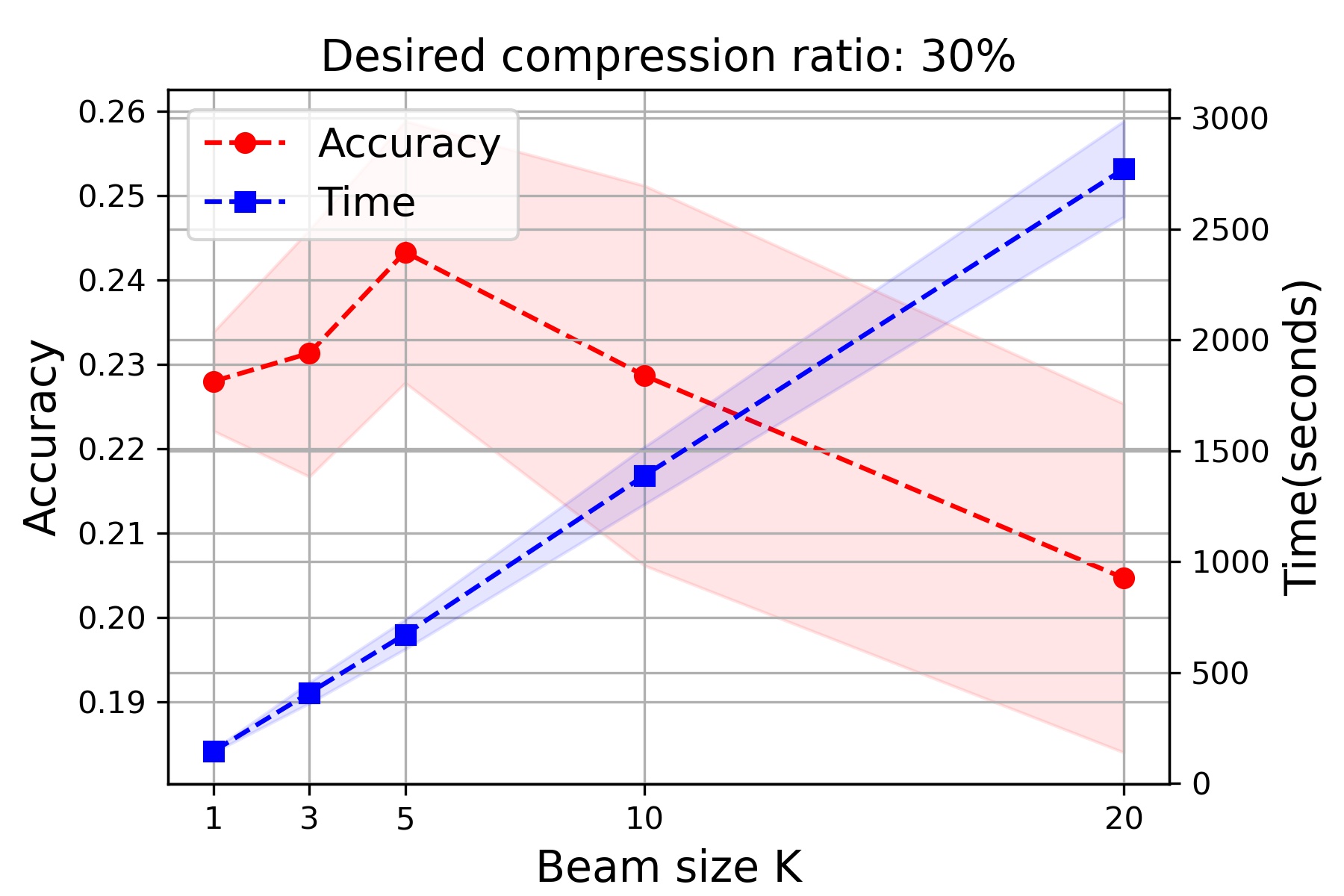}
        \caption{$s=20$}
        \label{subfig:s=20}
    \end{subfigure}
\caption{The effect of $K$ parameters on $\textit{mBS}$'s performance for a fixed $s$: a base neural network (ResNet56 on CIFAR-100) was truncated by the selected ranks and no further fine-tuning was applied for this analysis. Performance and search speed change as a function of $K$: (a) $s$ is fixed at 1, (b) $s$ is fixed at 3, (c) $s$ is fixed at 5, (d) $s$ is fixed at 10, (e) $s$ is fixed at 20.}
\label{fig:fixed s}
\end{figure}

\subsection{The effect of beam size $K$}
We checked the effect of the beam size $K$ for a fixed level step size $s$. As can be seen in \Cref{subfig:s=1}, \Cref{subfig:s=3}, \Cref{subfig:s=5}, \Cref{subfig:s=10}, we can confirm that the test accuracy improves as the beam size $K$ increases in general. This is because more candidates can be searched with a larger $K$. However, in \Cref{subfig:s=20} where $s$ is very large, it can be observed that the performance deteriorates as $K$ is increased. 

\newpage